\documentclass{article}

\PassOptionsToPackage{numbers, compress}{natbib}
\usepackage[final, nonatbib]{nips_2016}

\usepackage[utf8]{inputenc} % allow utf-8 input
\usepackage[T1]{fontenc}    % use 8-bit T1 fonts
\usepackage{hyperref}       % hyperlinks
\usepackage{url}            % simple URL typesetting
\usepackage{booktabs}       % professional-quality tables
\usepackage{amsfonts}       % blackboard math symbols
\usepackage{nicefrac}       % compact symbols for 1/2, etc.
\usepackage{microtype}      % microtypography
\usepackage[noadjust]{cite}

\DeclareMathAlphabet\mathbfcal{OMS}{cmsy}{b}{n}
\usepackage[tight,footnotesize]{subfigure}
\usepackage{amsthm}
\usepackage{bbm}
\usepackage{algorithm}
\usepackage{algorithmic}
\usepackage{graphicx,amsmath,amssymb,latexsym,epsfig}
\usepackage{setspace}
\usepackage{psfrag}

\usepackage{lscape}

\newtheorem*{proof of Theorem*}{Proof of Theorem 3}
\newtheorem{proof of Lemma}{Proof of Lemma}

\title{Nonnegative autoencoder with simplified random neural network}

\author{Yonghua Yin, Erol Gelenbe\\
Intelligent Systems and Networks Group,
Electrical \& Electronic Engineering Department\\
Imperial College, London SW7 2AZ, UK\\
 \texttt{y.yin14@imperial.ac.uk, e.gelenbe@imperial.ac.uk} \\
}

\begin{document}

\maketitle

\begin{abstract}
This paper proposes new nonnegative (shallow and multi-layer) autoencoders by combining the spiking Random Neural Network (RNN) model, the network architecture typical used in deep-learning area and the training technique inspired from nonnegative matrix factorization (NMF). The shallow autoencoder is a simplified RNN model, which is then stacked into a multi-layer architecture. The learning algorithm is based on the weight update rules in NMF, subject to the nonnegative probability constraints of the RNN. The autoencoders equipped with this learning algorithm are tested on typical image datasets including the MNIST, Yale face and CIFAR-10 datasets, and also using 16 real-world datasets from different areas. The results obtained through these tests yield the desired high learning and recognition accuracy. Also, numerical simulations of the stochastic spiking behavior of this RNN auto encoder, show that it can be implemented in a highly-distributed manner.
%This paper proposes new nonnegative (shallow and multi-layer) autoencoders by combining the model of spiking Random Neural Network (RNN), the network architecture in the deep-learning area and the training technique in the nonnegative matrix factorization (NMF) area. The shallow autoencoder is a simplified RNN model, which is then stacked into a multi-layer architecture. The learning algorithms are based on the weight update rules in the NMF area, subjecting to the nonnegative and probability constraints in a RNN model. The autoencoders equipped with the learning algorithms are tested on both typical image datasets including the MNIST, Yale face and CIFAR-10 datesets and 16 real-world datasets from different areas, and the results verify its efficacy. Simulation results of the stochastic spiking behaviors of this RNN autoencoder demonstrate that it can be implemented in a highly-distributed manner.
\end{abstract}

\section{Introduction}

A mathematical tool that has existed since 1989 \cite{RNN89,RNN90,RNN93}, but is
less well known in the machine-learning community, is the
Random Neural Network (RNN), which is a
stochastic
integer-state ``integrate and fire'' system
and developed to mimic the behaviour of biological neurons in the brain.
In an RNN, an arbitrarily large set of neurons interact with each other via excitatory
and inhibitory spikes which modify each neuron's action potential in
continuous time. The power of the RNN lays on the fact that, in steady state, the stochastic spiking behaviors of the network have a remarkable property
called ``product form'' and that the state probability distribution is given by an easily solvable system of non-linear equations. The RNN has been used for numerous applications \cite{Koubi,Cramer3,Hussain2,CPN2009,Fang,rubino2004new,mohamed2002study,radhakrishnan2011evaluating,ghalut2014non,stafylopatis1992pictorial,larijani2010voice,martinez2008grasp} that exploit its recurrent structure.

Deep learning has achieved great success in machine learning \cite{hinton2006reducing,lecun2015deep}.
Many computational models in deep learning exploit a feed-forward neural-network architecture that is composed of multi-processing layers, which allows the model to extract high-level representations from raw data. The feed-forward fully-connected multi-layer neural network could be difficult to train \cite{glorot2010understanding}. Pre-training the network layer by layer is a great advance \cite{hinton2006reducing,hinton2006fast} and useful due to its wide adaptability, though recent literature shows that utilizing the rectified linear unit (ReLU) could train a deep neural network without pre-training \cite{glorot2011deep}. The typical training procedure called stochastic gradient descent (SGD) provides a practical choice for handling large datasets \cite{bousquet2008tradeoffs}.

Nonnegative matrix factorization (NMF) is also a popular topic in machine learning \cite{lee1999learning,liu2010projective,hoyer2002non,wang2013nonnegative,ding2006orthogonal}, which learns part-based representations of raw data. Lee \cite{lee1999learning} suggested that the perception of the whole in the brain may be based on these part-based representations (based on the physiological evidence \cite{wachsmuth1994recognition}) and proposed simple yet effective update rules.
Hoyer \cite{hoyer2002non} combined sparse coding and NMF that allows control over sparseness.
Ding investigated the equivalence between the NMF and K-means clustering in \cite{ding2005equivalence,ding2006orthogonal} and presented simple update rules for orthogonal NMF.
Wang \cite{wang2013nonnegative} provided a comprehensive review on recent processes in the NMF area.

This paper first exploits the structure of the RNN equations as a quasi-linear structure. Using it in the feed-forward case, an RNN-based shallow nonnegative autoencoder is constructed. Then, this shallow autoencoder is stacked into a multi-layer feed-forward autoencoder following the network architecture in the deep learning area \cite{hinton2006reducing,hinton2006fast,lecun2015deep}. Since connecting weights in the RNN are products of firing rates and transition probabilities, they are subject to the constraints of nonnegativity and that the sum of probabilities is no larger than 1, which are called the RNN constraints in this paper.
In view of that, the conventional gradient descent is not applicable for training such an autoencoder.
By adapting the update rules from nonnegative graph embedding that can be seemed as a variant of NMF, applicable update rules are developed for the autoencoder that satisfy the first RNN constraint of nonnegativity. For the second RNN constraint, we impose a check-and-adjust procedure into the iterative learning process of the learning algorithms. The training procedure of SGD is also adapted into the algorithms. The efficacy of the nonnegative autoencoders equipped with the learning algorithms is well verified via numerical experiments on both typical image datasets including the MNIST \cite{lecun1998gradient}, Yale face \cite{cai2007learning} and CIFAR-10 \cite{krizhevsky2009learning} datesets and 16 real-world datasets in different areas from the UCI machine learning repository \cite{Lichman:2013}. Then, we simulate the spiking behaviors of the RNN-based autoencoder, where simulation results conform well with the corresponding numerical results, therefore demonstrating that this nonnegative autoencoder can be implemented in a highly-distributed and parallel manner.

\section{A quasi-linear simplified random neural network}

An arbitrary neuron in the RNN can receive excitatory or inhibitory spikes from external sources, in which case they arrive according to independent  Poisson processes.
Excitatory or inhibitory spikes can also arrive from other neurons to a given neuron, in which case they arrive when the sending neuron fires,
which happens only if that neuron's input state is positive (i.e. the neuron is excited) and inter-firing intervals from
the same neuron $v$ are exponentially distributed random variables with rate $r_v\geq 0$. Since the firing times depend on the internal state of the sensing neuron, the
arrival process of neurons from other cells is not in general Poisson. From the preceding assumptions it was proved in \cite{RNN93} that for an arbitrary $N$ neuron RNN, which may or may not be recurrent
(i.e. containing feedback loops), the probability in steady-state that any cell $h$, located anywhere in the network, is excited is given by the expression:
\begin{equation}
q_h=\min(\frac{\lambda_h^++\sum_{v=1}^Nq_{v}r_vp^+_{vh}}{r_h+\lambda_h^-+\sum_{v=1}^Nq_{v}r_vp^-_{vh}},1), \label{RNN}
\end{equation}
for $h=1,~...~,N$, where $p^+_{vh},~p^-_{vh}$ are the probabilities that cell $v$ may send excitatory or inhibitory spikes to cell $h$, and $\lambda^+_h,~\lambda^-_h$ are the external arrival rates of excitatory and inhibitory spikes to
neuron $h$. Note that $\min(a,b)$ is a element-wise operation whose output is the smaller one between $a$ and $b$. In \cite{RNN93}, it was shown that the system of $N$ non-linear equations (\ref{RNN}) have a solution which is unique.

%The formula (\ref{RNN}) lends itself to various approximations such as the first order approximation:
%\begin{equation}
%q_h\approx\frac{\lambda_h^++\sum_{v=1}^Nq_{v}r_vp^+_{vh}}{r_h}[1-\frac{\lambda_h^-+\sum_{v=1}^Nq_{v}r_vp^-_{vh}}{r_h}], \label{apRNN}
%\end{equation}
%provided that $r_h>>\lambda_h^-+\sum_{v=1}^Nq_{v}r_vp^-_{vh}$.

Before adapting the RNN as a non-negative autoencoder (Section \ref{sec.shallow}), we will simplify the recurrent RNN model into the feed-forward structure shown in Figure \ref{fig.SSM}. The simplified RNN has an input layer and a hidden layer. The $V$ input neurons receive excitatory spikes from the outside world, and they fire excitatory spikes to the $H$ hidden neurons.

Let us denote by  $\hat{q}_v$ the probability that the $v$th input neuron  ($v=1,\cdots,V$) is excited and $q_h$ the probability that the $h$th hidden neuron ($h=1,\cdots,H$) is excited. According to \cite{RNN89} and (\ref{RNN}), they are given by
%$\hat{q}_v = \min({\hat{\Lambda}^+_v}/{\hat{r}_v},1)$ and $q_h = \min({\Lambda^+_h}/{r_h},1)$
$\hat{q}_v = \min({\hat{\Lambda}^+_v}/{\hat{r}_v},1)$, and $q_h = \min({\Lambda^+_h}/{r_h},1)$,
where the quantities $\hat{\Lambda}^+_v$ and $\Lambda^+_h$ represent the total average arrival rates of excitatory spikes, $\hat{r}_v$ and $r_h$ represent the firing rates of the neurons.
Neurons in this model interact with each other in the following manner, where $h=1,\cdots,H$ and $v=1,\cdots,V$.
When the $v$th input neuron fires, it sends excitatory spikes to the $h$th hidden neuron with probability $p^{+}_{v,h}\geq 0$. Clearly, $\sum_{h=1}^{H}p^{+}_{v,h} \leq 1$.\\
$\bullet$ The $v$th input neuron receives excitatory spikes from the outside world with rate $x_v \geq 0$. \\
$\bullet$ When the $h$th hidden neuron fires, it sends excitatory spikes outside the network.

%\begin{itemize}
%\item When the $v$th input neuron fires, it sends excitatory spikes to the $h$th hidden neuron with probability $p^{+}_{v,h}\geq 0$. Clearly, $\sum_{h=1}^{H}p^{+}_{v,h} \leq 1$.
%\item The $v$th input neuron receives excitatory spikes from the outside world with rate $x_v \geq 0$.
%%\item  The firing rate of the $h$th hidden neuron $r_h=1$.
%\item When the $h$th hidden neuron fires, it sends excitatory spikes outside the network.
%\end{itemize}
Let us denote $w_{v,h} = p^{+}_{v,h} \hat{r}_v$.
For simplicity,
let us set the firing rates of all neurons to $\hat{r}_v=r_h=1$ or that $\sum_{h=1}^{H}w_{v,h} \leq 1$.
Then, $\hat{\Lambda}^+_v=x_v$, $\hat{r}_v=1$, $\Lambda^+_h= \sum_{v=1}^{V}w_{v,h} \hat{q}_v$, and
using the fact that $q_h,~q_v$ are probabilities, we can write:
\begin{eqnarray} \label{eqn.srnn}
\hat{q}_v = \min(x_v,1), ~q_h = \min(\sum_{v=1}^{V}w_{v,h} \hat{q}_v,1),
\end{eqnarray}
subject to $\sum_{h=1}^{H}w_{v,h} \leq 1$.
We can see from (\ref{eqn.srnn}) that this simplified RNN is quasi linear. For the network shown in Figure \ref{fig.SSM}, we call it a quasi-linear RNN (LRNN).

%Let a $V$-vector $\hat{Q}=[\hat{q}_v]$, a $V$-vector $X=[x_v]$
%a $H$-vector $Q=[q_h]$ and a $V \times U$-matrix $W=[w_{v,h}]$.
%Then, (\ref{eqn.srnn1}) could be rewritten as
%\begin{equation} \label{eqn.srnn1matrix}
%\begin{split}
%&\hat{Q} = \min(X,1),\\
%& Q = \min(\hat{Q} W ,1),\\
%\end{split}
%\end{equation}
%subject to $\sum_{h=1}^{H}w_{v,h} \leq 1$.
%We can see from (\ref{eqn.srnn1matrix}) that this simplified RNN is nearly linear. For the network shown in Figure \ref{fig.SSM}, we could call it a nearly-linear RNN (LRNN).

\section{Shallow non-negative LRNN autoencoder}\label{sec.shallow}

\begin{figure}[t]
\centering
\psfrag{W}[c][c][0.8]{$w_{v,h}^{+}$}%
\psfrag{qc}[c][c][0.8]{$q_{h}$}%
\psfrag{dot}[c][c][0.8]{$\cdots$}%
\psfrag{qu}[c][c][0.8]{$\hat{q}_v$}%
\psfrag{cluster}[c][c][0.8]{~~~Hidden}%
\psfrag{input}[c][c][0.8]{~~~~Input}%
\includegraphics[width=4.3in]{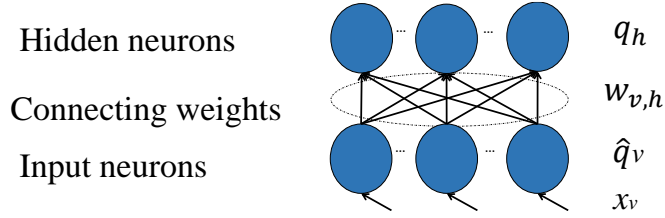}
\caption{Brief model structure of the quasi-linear RNN.}
\label{fig.SSM}
\end{figure}

We add an output layer with $O$ neurons on top of the hidden layer of the LRNN shown in Figure \ref{fig.SSM} to construct a shallow non-negative LRNN autoencoder.
Let $\overline{q}_o$ denote the probability that the $o$th output neuron is excited, and
the $o$th output neurons interact with the LRNN in the following manner, where $o=1,\cdots,O$.\\
$\bullet$ When the $h$th hidden neuron fires, it sends excitatory spikes to the $o$th output neuron with probability $\overline{p}^{+}_{h,o}\geq 0$. Also, $\sum_{o=1}^{O}\overline{p}^{+}_{h,o} \leq 1$.\\
$\bullet$ The firing rate of the $o$th output neuron $\overline{r}_o=1$.\\
Let $\overline{w}_{h,o}=\overline{p}^{+}_{h,o}r_h=\overline{p}^{+}_{h,o}$. Then, $\sum_{o=1}^{O}\overline{w}_{h,o} \leq 1$. The shallow LRNN autoencoder is described by
\begin{eqnarray} \label{eqn.srnn1}
\hat{q}_v = \min(x_v,1), ~
q_h = \min(\sum_{v=1}^{V}w_{v,h} \hat{q}_v,1),~
\overline{q}_o = \min(\sum_{h=1}^{H} \overline{w}_{h,o} q_h,1),
\end{eqnarray}
where $O=V$ and the input, hidden and output layers are the visual, encoding and decoding layers.

Suppose there is a dataset represented by a nonnegative $D \times V$ matrix $X=[x_{d,v}]$, where $D$ is the number of instances, each instance has $V$ attributes and $x_{d,v}$ is the $v$th attribute of the $d$th instance. We import $X$ into the input layer of the LRNN autoencoder. Let $\hat{q}_{d,v}$, $q_{d,h}$ and $\overline{q}_{d,o}$ respectively denote the values of $\hat{q}_{v}$, $q_{d}$ and $\overline{q}_{o}$ for the $d$th instance.

Let a $D \times V$-matrix $\hat{Q}=[\hat{q}_{d,v}]$,
a $D \times H$-matrix $Q=[q_{d,h}]$, a $D \times O$-matrix $\overline{Q}=[\overline{q}_{d,o}]$, a $V \times U$-matrix $W=[w_{v,h}]$ and a $H \times O$-matrix $\overline{W}=[\overline{w}_{h,o}]$.
Then, (\ref{eqn.srnn1}) can be rewritten as the following matrix manner:
\begin{equation} \label{eqn.srnn1matrix}
\hat{Q} = \min(X,1),~ Q = \min(\hat{Q} W ,1), ~ \overline{Q} = \min(Q \overline{W},1),
\end{equation}
subject to the RNN constraints $W \geq 0$, $\overline{W} \geq 0$, $\sum_{h=1}^{H}w_{v,h} \leq 1$ and $\sum_{o=1}^{O}\overline{w}_{h,o} \leq 1$.
The problem for the autoecoder to learn the dataset $X$
can be described as
\begin{equation} \label{eqn.problem1}
\begin{split}
&\arg \min_{W, \overline{W}} ||X - \overline{Q}||^2, ~ \text{s.t.~}W \geq 0, \overline{W} \geq 0, \sum_{h=1}^{H}w_{v,h} \leq 1, \sum_{o=1}^{O}\overline{w}_{h,o} \leq 1.
\end{split}
\end{equation}

We use the following update rules to solve this problem, which are simplified from Liu's work \cite{NNPCA}:
\begin{equation} \label{eqn.w1_update}
\begin{split}
&w_{v,h} \leftarrow w_{v,h}\frac{(X^{\text{T}} X \overline{W}^{\text{T}})_{v,h}}{ (X^{\text{T}} X W \overline{W} \overline{W}^{\text{T}})_{v,h}},
\end{split}
\end{equation}
\begin{equation} \label{eqn.w2_update}
\begin{split}
&\overline{w}_{h,o} \leftarrow \overline{w}_{h,o}\frac{(W^{\text{T}}X^{\text{T}} X)_{h,o} }{ (W^{\text{T}}X^{\text{T}} X W \overline{W})_{h,o}},
\end{split}
\end{equation}
where the symbol $(\cdot)_{v,h}$ denotes the element in the $v$th row and $h$th column of a matrix.
Note that, to avoid the division-by-zero problem, zero elements in the denominators of (\ref{eqn.w1_update}) and (\ref{eqn.w2_update}) are replaced with tiny positive values, (e.g., ``eps'' in MATLAB).
After each update, adjustments need to be made such that $W$ and $\overline{W}$ satisfy the RNN constraints. The procedure to train the shallow LRNN autoencoder (\ref{eqn.srnn1matrix}) is given in Algorithm \ref{algorithm_lrnn}, where the operation $\max(W)$ produces the maximal element in $W$,
the operations of $w_{v,h}\leftarrow w_{v,h}/\sum_{h=1}^{H}w_{v,h}$ and $\overline{w}_{h,o}\leftarrow \overline{w}_{h,o}/\sum_{o=1}^{O}\overline{w}_{h,o}$ guarantee that the weights satisfy the RNN constraints, and the operations $W \leftarrow W/\max(\bar{X}W)$ and $\overline{W} \leftarrow \overline{W}/\max(H\overline{W})$ normalize the weights to reduce the number of neurons that are saturated.

\begin{algorithm}
\caption{Procedure for training a shallow nonnegatvie LRNN autoencder (\ref{eqn.srnn1matrix}) }\label{algorithm_lrnn}
\begin{algorithmic}
\STATE Randomly initialize $W$ and $\overline{W}$ that satisfy RNN constraints
\STATE \textbf{while} terminal condition is not satisfied {do}
\STATE ~~~~~\textbf{for} each minibatch $\bar{X}$ \textbf{do}
\STATE ~~~~~~~~~~update $W$ with (\ref{eqn.w1_update})
\STATE ~~~~~~~~~~\textbf{for} $v=1,\cdots,V$ \textbf{do}
\STATE ~~~~~~~~~~~~~~~~~~~~\textbf{if} $\sum_{h=1}^{H}w_{v,h} >1$
\STATE ~~~~~~~~~~~~~~~~~~~~~~~~~$w_{v,h}\leftarrow w_{v,h}/\sum_{h=1}^{H}w_{v,h}, \text{for} ~ h=1,\cdots,H$
\STATE ~~~~~~~~~~$W \leftarrow W/\max(\bar{X}W)$
\STATE ~~~~~~~~~~update $\overline{W}$ with (\ref{eqn.w2_update})
\STATE ~~~~~~~~~~\textbf{for} $h=1,\cdots,H$ \textbf{do}
\STATE ~~~~~~~~~~~~~~~~~~~~\textbf{if} $\sum_{o=1}^{O}\overline{w}_{h,o} >1$
\STATE ~~~~~~~~~~~~~~~~~~~~~~~~~$\overline{w}_{h,o}\leftarrow \overline{w}_{h,o}/\sum_{o=1}^{O}\overline{w}_{h,o}, \text{for} ~ o=1,\cdots,O$
\STATE ~~~~~~~~~~$H=\min(\bar{X}W,1)$
\STATE ~~~~~~~~~~$\overline{W} \leftarrow \overline{W}/\max(H\overline{W})$
\end{algorithmic}
\end{algorithm}

\section{Multi-layer non-negative LRNN autoencoder}

We stack multi LRNNs to build a multi-layer non-negative LRNN autoencoder.
Suppose the multi-layer autoencoder has a visual layer, $M$ encoding layers and $M$ decoding layer ($M \geq 2$), and they are connected in series with excitatory weights $W_m$ and $\overline{W}$ with $m=1,\cdots,M$. We import a dataset $X$ into the visual layer of the autoencoder.
Let $H_m$ and $O_m$ denote the numbers of neurons in the $m$th encoding layer and decoding layer, respectively.
For the autoencoder, $V=O_M$, $H_m=O_{M-m}$ with $m=1,\cdots,M-1$.

Let $\hat{Q}$ denote the state of the visual layer, $Q_m$ denote the state of the $m$th encoding layer and $\overline{Q}_m$ denote the state of the $m$th decoding layer. Then, the multi-layer LRNN autoencoder is described by
\begin{equation} \label{eqn.multi}
\begin{cases}
& \hat{Q} = \min(X,1),~
 Q_1 = \min(\hat{Q} W_1,1),~
 Q_m = \min(Q_{m-1} W_m,1),\\
& \overline{Q}_1 = \min(Q_M \overline{W}_1,1),~
 \overline{Q}_m = \min(\overline{Q}_{m-1} \overline{W}_m,1),\\
\end{cases}
\end{equation}
with $m=2,\cdots,M$. The RNN constraints for (\ref{eqn.multi}) are
$W_m \geq 0$, $\overline{W}_m \geq 0$ and the summation of each row in $W_m$ and $\overline{W}_m$ is not larger than 1, where $m=1,\cdots,M$.
The problem for the multi-layer LRNN autoencoder (\ref{eqn.multi}) to learn dataset $X$ can be described as
\begin{equation} \label{eqn.problem1}
\begin{split}
&\arg \min_{W_m, \overline{W}_m} ||X - \overline{Q}_M||^2,\\
%&\text{s.t.~}W_m \geq 0, \overline{W}_m \geq 0,
\end{split}
\end{equation}
subject to the RNN constraints, where $m=1,\cdots,M$.
The procedure to train the multi-layer non-negative LRNN autoencder (\ref{eqn.multi}) is given in Algorithm \ref{algorithm_multi}.

To avoid loading the whole dataset into the computer memory, we could also use Algorithm \ref{algorithm_multi2} to train the autoencoder, where the update rules could be
\begin{equation} \label{eqn.w1_update2}
\begin{split}
&W_1 \leftarrow W_1 \odot \frac{\hat{Q}^{\text{T}} \hat{Q} \overline{W}_M^{\text{T}}}{\hat{Q}^{\text{T}} \hat{Q} W_1 \overline{W}_M \overline{W}_M^{\text{T}}},~
W_m \leftarrow W_m \odot \frac{Q_{m-1}^{\text{T}} Q_{m-1} \overline{W}_{M-m+1}^{\text{T}}}{ Q_{m-1}^{\text{T}} Q_{m-1} W_m \overline{W}_{M-m+1} \overline{W}_{M-m+1}^{\text{T}}},
\end{split}
\end{equation}
\begin{equation} \label{eqn.w2_update2}
\begin{split}
&\overline{W}_M \leftarrow \overline{W}_M \odot \frac{W_1^{\text{T}}\hat{Q}^{\text{T}} \hat{Q}}{W_1^{\text{T}}\hat{Q}^{\text{T}} \hat{Q} W_1 \overline{W}_M},~
\overline{W}_{M-m+1} \leftarrow \overline{W}_{M-m+1} \odot \frac{W_m^{\text{T}}Q_{m-1}^{\text{T}} Q_{m-1} }{W_m^{\text{T}}Q_{m-1}^{\text{T}} Q_{m-1} W_m \overline{W}_{M-m+1}},
\end{split}
\end{equation}
with $m=2,\cdots,M$ and the operation $\odot$ denoting element-wise product of two matrices.
To avoid the division-by-zero problem, zero elements in denominators of (\ref{eqn.w1_update2}) and (\ref{eqn.w2_update2}) are replaced with tiny positive values.
The operations of adjusting the weights to satisfy the RNN constraints and normalizing the weights are the same as those in Algorithm \ref{algorithm_lrnn}.

\begin{algorithm}
\caption{Proceduce for training a multi-layer LRNN-based non-negatvie autoencder (\ref{eqn.multi}) }\label{algorithm_multi}
\begin{algorithmic}
\STATE $X_1=X$
\STATE \textbf{for} $m=1,\cdots,M$ \textbf{do}
\STATE ~~~~~Train $W_m$ and $\overline{W}_{M-m+1}$ with Algorithm \ref{algorithm_lrnn} that takes $X_m$ as input dataset
\STATE ~~~~~\textbf{if} $m \neq M$ \textbf{do}
\STATE ~~~~~~~~~~$X_{m+1}=\min(X_m W_m,1)$
\end{algorithmic}
\end{algorithm}

\begin{algorithm}
\caption{Proceduce for training a multi-layer LRNN-based non-negatvie autoencder (\ref{eqn.multi}) (minibatch manner)}\label{algorithm_multi2}
\begin{algorithmic}
\STATE Randomly initialize $W_m$ and $\overline{W}_m$ that satisfy RNN constraints (with $m=1,\cdots,M$)
\STATE \textbf{while} terminal condition is not satisfied {do}
\STATE ~~~~~\textbf{for} each minibatch $\bar{X}$ \textbf{do}
\STATE ~~~~~~~~~~\textbf{for} $m=1,\cdots,M$ \textbf{do}
\STATE ~~~~~~~~~~~~~~~update $W_m$ with (\ref{eqn.w1_update2})
\STATE ~~~~~~~~~~~~~~~adjust $W_m$ to satisfy RNN constraints
\STATE ~~~~~~~~~~~~~~~normalize $W_m$ subject to $\bar{X}$
\STATE ~~~~~~~~~~~~~~~update $\overline{W}_m$ with (\ref{eqn.w2_update2})
\STATE ~~~~~~~~~~~~~~~adjust $\overline{W}_m$ to satisfy RNN constraints
\STATE ~~~~~~~~~~~~~~~normalize $\overline{W}_m$ subject to $\bar{X}$
\end{algorithmic}
\end{algorithm}

%\begin{algorithm}
%\caption{Proceduce for training a multi-layer LRNN-based non-negatvie autoencder (\ref{eqn.multi}) }\label{algorithm_multi}
%\begin{algorithmic}
%\STATE Pre-train the autoencoder layer-wise with Algorithm \ref{algorithm_lrnn}
%\STATE \textbf{for} each minibatch \textbf{do}
%\STATE ~~\textbf{for} $m=1,\cdots,M$ \textbf{do}
%\STATE ~~~~update $W_m$ with (\ref{eqn.update_whole1})
%\STATE ~~~~adjust $W_m$ to satisfy the RNN constraints
%\STATE ~~~~update $\overline{W}_m$ with (\ref{eqn.update_whole2})
%\STATE ~~~~adjust $\overline{W}_m$ to satisfy the RNN constraints
%\STATE ~~\textbf{end for}
%\STATE \textbf{end for}
%\end{algorithmic}
%\end{algorithm}

\section{Numerical Experiments}

\subsection{Datasets}

{\bf MNIST:} The MNIST dataset of handwritten digits \cite{lecun1998gradient} contains 60,000 and 10,000 images in the training and test dataset. The number of input attributes is 784 ($28 \times 28$ images), which are in $[0,1]$.
%The number of classes is 10.

{\bf Yale face:} This database (\url{http://vision.ucsd.edu/content/yale-face-database}) contains 165 gray scale
images of 15 individuals.
Here we use the pre-processed dataset from \cite{cai2007learning}, where each image is resized as $32 \times 32$ (1024 pixels).

{\bf CIFAR-10:}
The CIFAR-10 dataset consists of 60,000 $32 \times 32$ colour images \cite{krizhevsky2009learning}. Each image has 3072 attributes. It contains 50,000 and 10,000 images in the training and test dataset.

\begin{table}[h] \centering
\setlength{\belowcaptionskip}{5pt}
  \caption{Features of different UCI real-world datasets from different areas}
  \begin{tabular} {|l|c|c|}
    \hline
  Dataset  &Attr. No. &Inst. No. \\
  \hline
{\it Iris}&4&150 \\ \hline
{\it Teaching Assistant Evaluation (TAE)} &5 &151\\\hline
{\it Liver Disorders (LD)}&5  &345\\\hline
{\it Seeds} &7 &210\\\hline
{\it Pima Indians Diabetes (PID)}&8  &768\\\hline
%{\it Yeast \cite{horton1996probabilistic}}&8  &1484\\\hline
{\it Breast Cancer Wisconsin (BC) \cite{wolberg1990multisurface,mangasarian1990pattern,bennett1992robust}} &9 &699 \\\hline
{\it Glass }&9 &214 \\\hline
{\it Wine}&13 &178 \\\hline
{\it Zoo } &16 &100 \\\hline
{\it Parkinsons \cite{little2007exploiting}} &22 &195\\\hline
{\it Wall-Following Robot Navigation (WFRN) \cite{freire2009short}} &24 &5456 \\\hline
{\it Ionosphere \cite{sigillito1989classification}}&34 &351 \\\hline
{\it Soybean Large (SL)}&35 &186\\\hline
{\it First-Order Theorem Proving (FOTP) \cite{bridge2014machine}} &51 &6118\\\hline
{\it Sonar \cite{gorman1988analysis}} &60 &208\\\hline
{\it Cardiac Arrhythmia (CA) \cite{guvenir1997supervised}} &279 &452\\
\hline
  \end{tabular} \label{tab.ucidataset}
\end{table}

{\bf UCI real-world datasets:} In addition to image datasets, we also conduct numerical experiments on different real-world datasets in different areas from the UCI machine learning repository \cite{Lichman:2013}. The names, attribute numbers and instance numbers of these datasets  are listed in Table \ref{tab.ucidataset}.

\subsection{Convergence and reconstruction performance}

%Let us first test the reconstruction performance of the shallow non-negative LRNN autoencoder. We use different structures of $784-500$, $784-300$, $784-100$, $784-50$,  $784-10$ and the MNIST dataset for experiments. The whole training dataset of 60,000 images is use for training. Figures \ref{fig.mnist_shallow} shows the curves of training error (mean square error) versus the number of iterations, where, in each iteration, a minibatch of size 100 is handled.

\begin{figure}\centering
\subfigure[MNIST]{\label{fig.mnist}\includegraphics[width=1.7in]{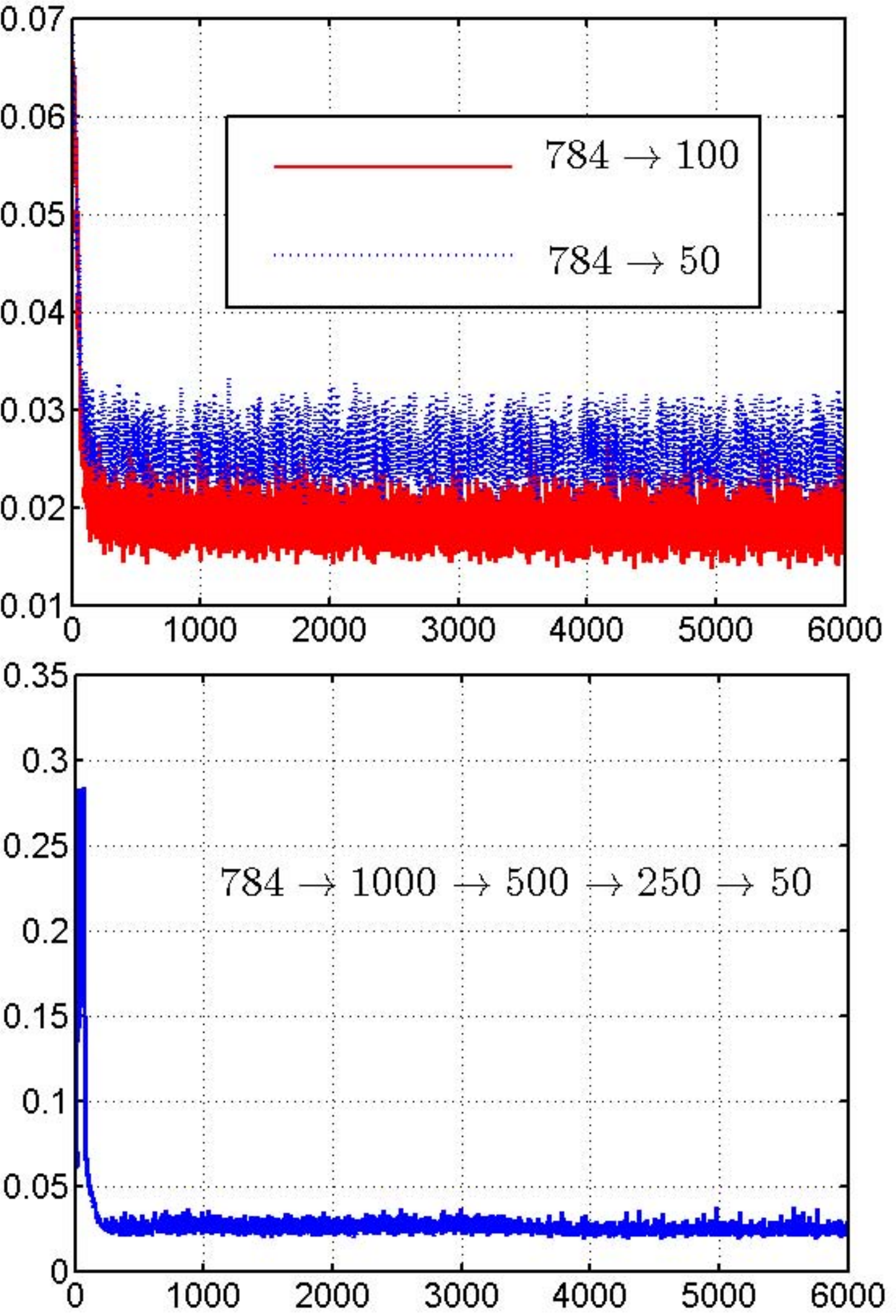}}
\subfigure[Yale face]{\label{fig.yale}\includegraphics[width=1.7in]{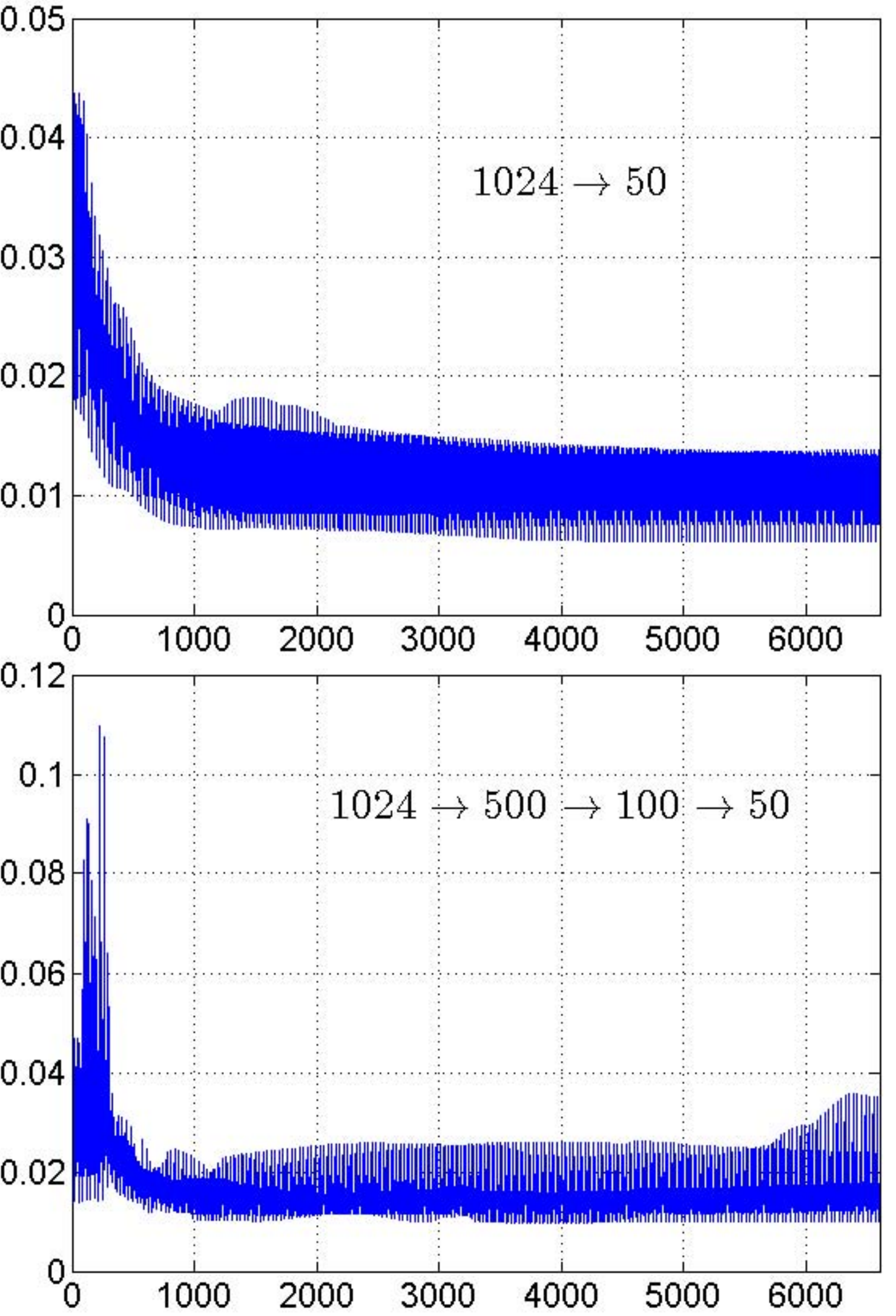}}
\subfigure[CIFAR-10]{\label{fig.cifar}\includegraphics[width=1.7in]{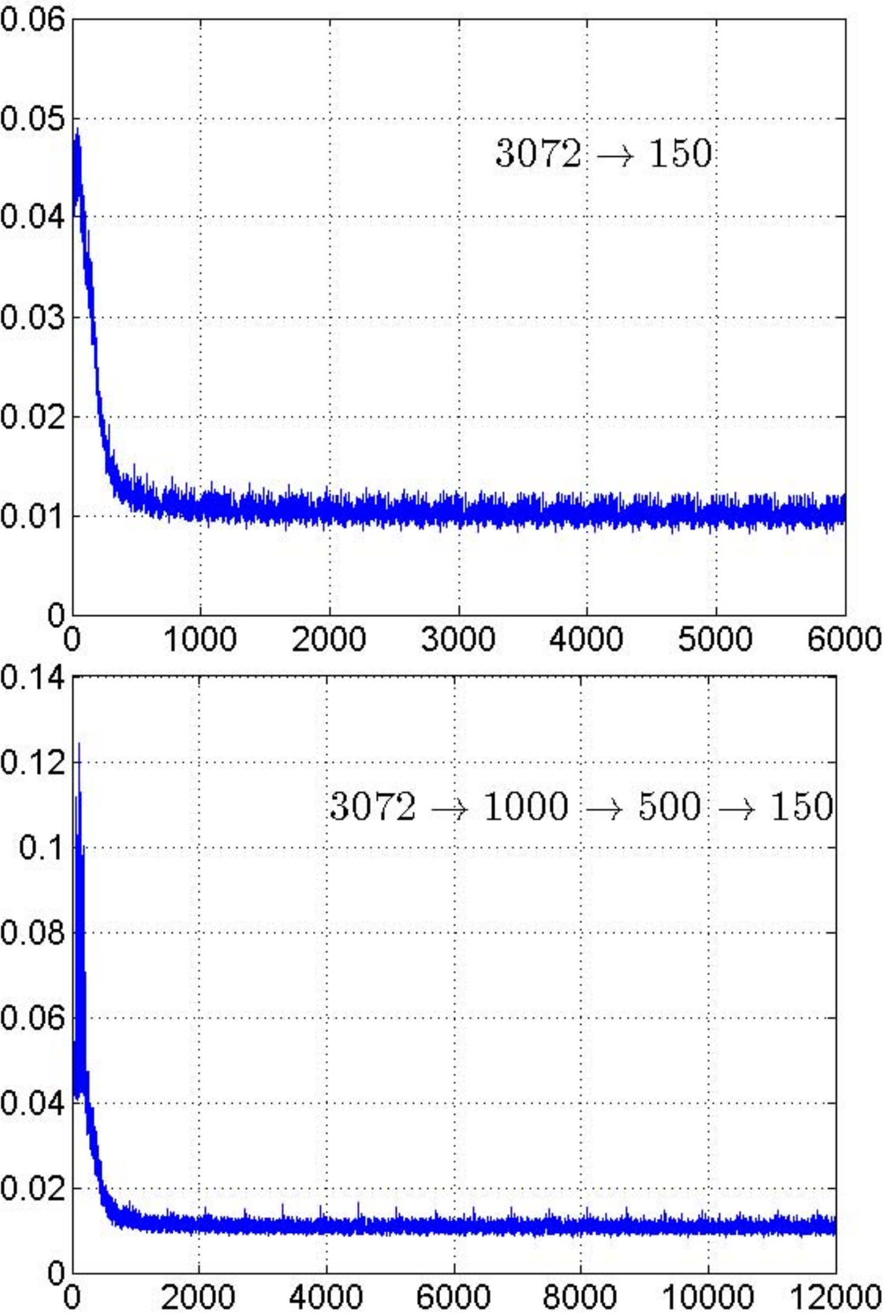}}
\caption{Reconstruction error (Y-axis) versus iteration number (X-axis) of shallow and multi-layer LRNN autoencoders for the MNIST, Yale face and CIFAR-10 datasets.} \label{fig.image}
\end{figure}

{\bf Results of MNIST:} \label{sec.mnist_results}
Let us first test the convergence and reconstruction performance of the shallow non-negative LRNN autoencoder. We use structures of $784 \rightarrow 100$ (for simplicity, we use the encoding part to represent an autoencoder) and $784 \rightarrow 50$ and the MNIST dataset for experiments. The whole training dataset of 60,000 images is used for training. Figure \ref{fig.mnist} shows the curves of training error (mean square error) versus the number of iterations, where, in each iteration, a minibatch of size 100 is handled.
Then, we use a multi-layer non-negative LRNN autoencoder with structure $784\rightarrow 1000 \rightarrow 500 \rightarrow 250 \rightarrow 50$, and
the corresponding curve of training error versus iterations is also given in Figure \ref{fig.mnist}.
It can be seen from Figure \ref{fig.mnist} that reconstruction errors using the LRNN autoencoders equipped with the developed algorithms converge well for different structures. In addition, the lowest errors using the shallow and multi–layer autoencoders are respectively 0.0204 and 0.0190. The results show that, for the same encoding dimension, the performances of the shallow and multi-layer structures are similar for this dataset.

{\bf Results of Yale face:} Attribute values are normalized into $[0,1]$ (by dividing by 255). The structures for the shallow and multi-layer LRNN autoencoders are respectively $1024 \rightarrow 50$ and $1024 \rightarrow 500 \rightarrow 100 \rightarrow 50$. The size of a minibatch is 5. Curves of reconstruction errors versus iterations are given in Figure \ref{fig.yale}.
For this dataset, the shallow autoencoder seems more stable than the multi-layer one.

{\bf Results of CIFAR-10:} Attribute values of the dataset are also divided by 255 for normalization in range $[0,1]$. The structures used are $3072\rightarrow 150$ and $3072 \rightarrow 1000 \rightarrow 500 \rightarrow 150$. Both the training and testing dataset (total 60,000 images) are used for training the autoencoders. The size of minibatch is chosen as 100. The results are given in Figure \ref{fig.cifar}. We can see that reconstruction errors for both structures converge as the number of iterations increases. In addition, the lowest reconstruction errors in using the shallow and multi-layer autoencoders are the same (0.0082). These results together with those with the MNIST and Yale face datasets (Figures \ref{fig.mnist} to \ref{fig.cifar}) verify the good convergence and reconstruction performance of both the shallow and multi-layer LRNN autoencoders for handling image datsets.

\begin{figure*}[t]\centering
\psfrag{iteration}[c][c][0.6]{Iterations}%
\psfrag{mse}[c][c][0.6]{Reconstruction error}%
\subfigure[IRIS]{\label{fig.iris}\includegraphics[width=1.3in]{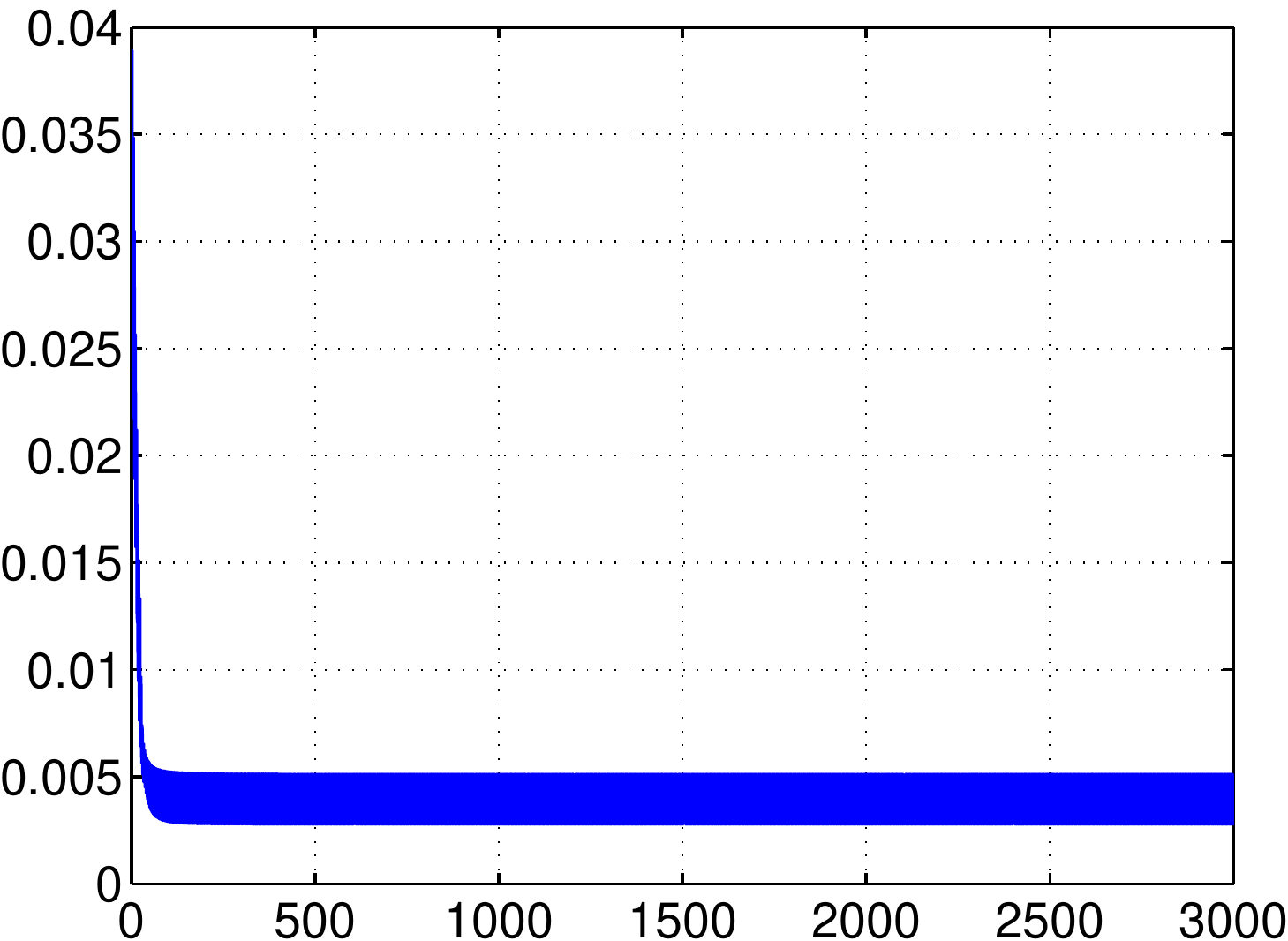}}
\subfigure[TAE]{\label{fig.tae}\includegraphics[width=1.3in]{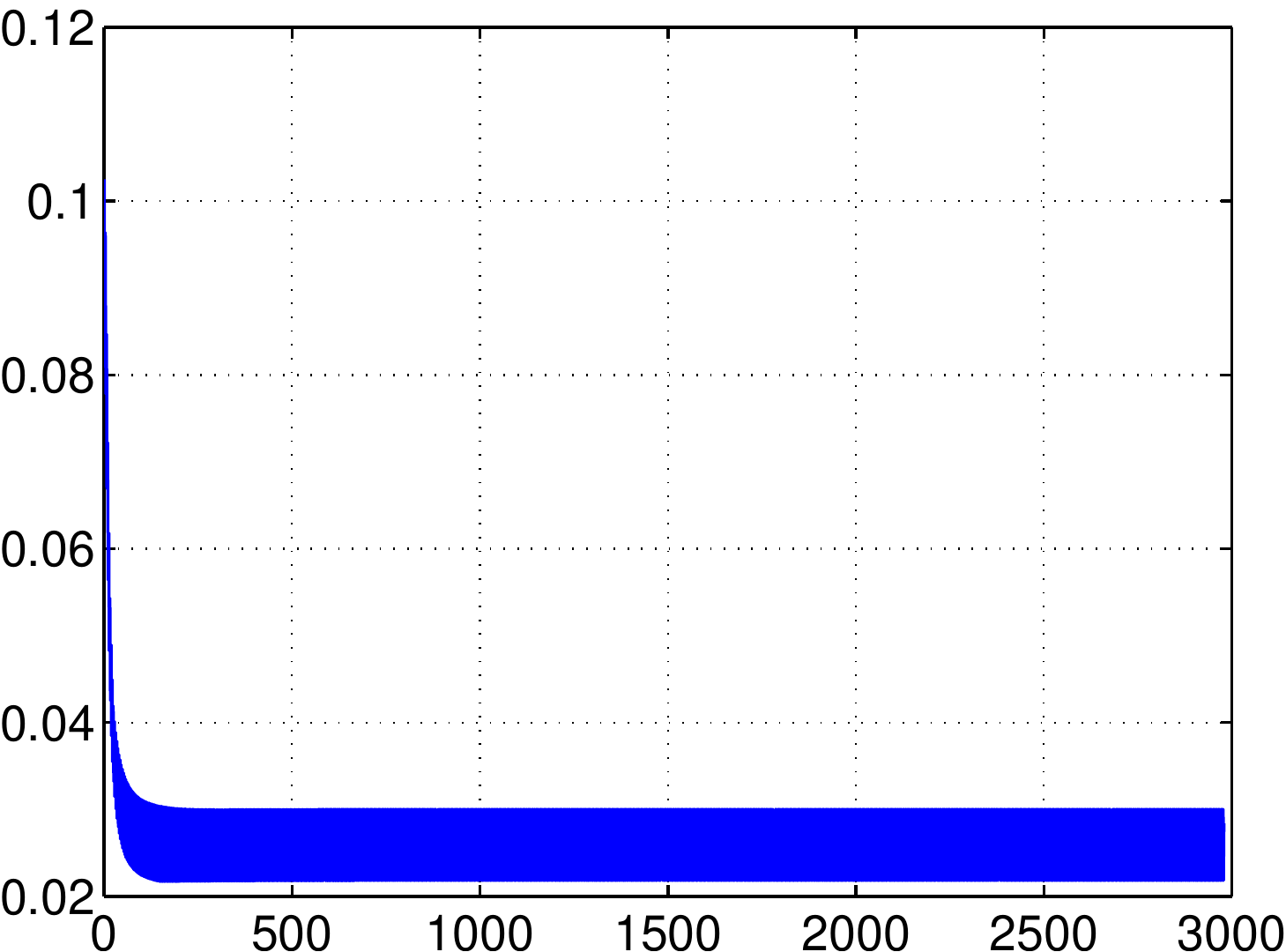}}
\subfigure[LD]{\label{fig.iris}\includegraphics[width=1.3in]{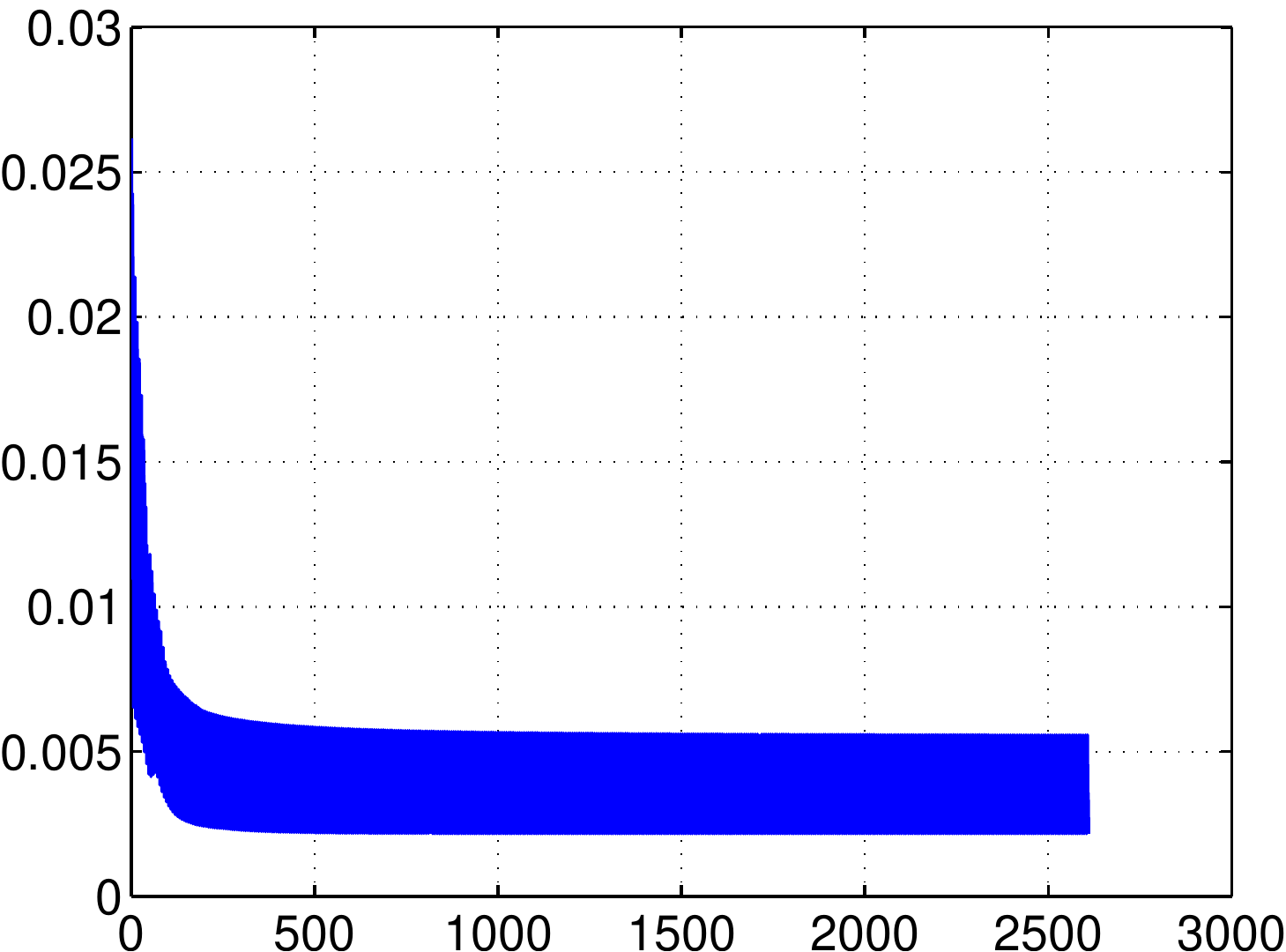}}
\subfigure[Seeds]{\label{fig.tae}\includegraphics[width=1.3in]{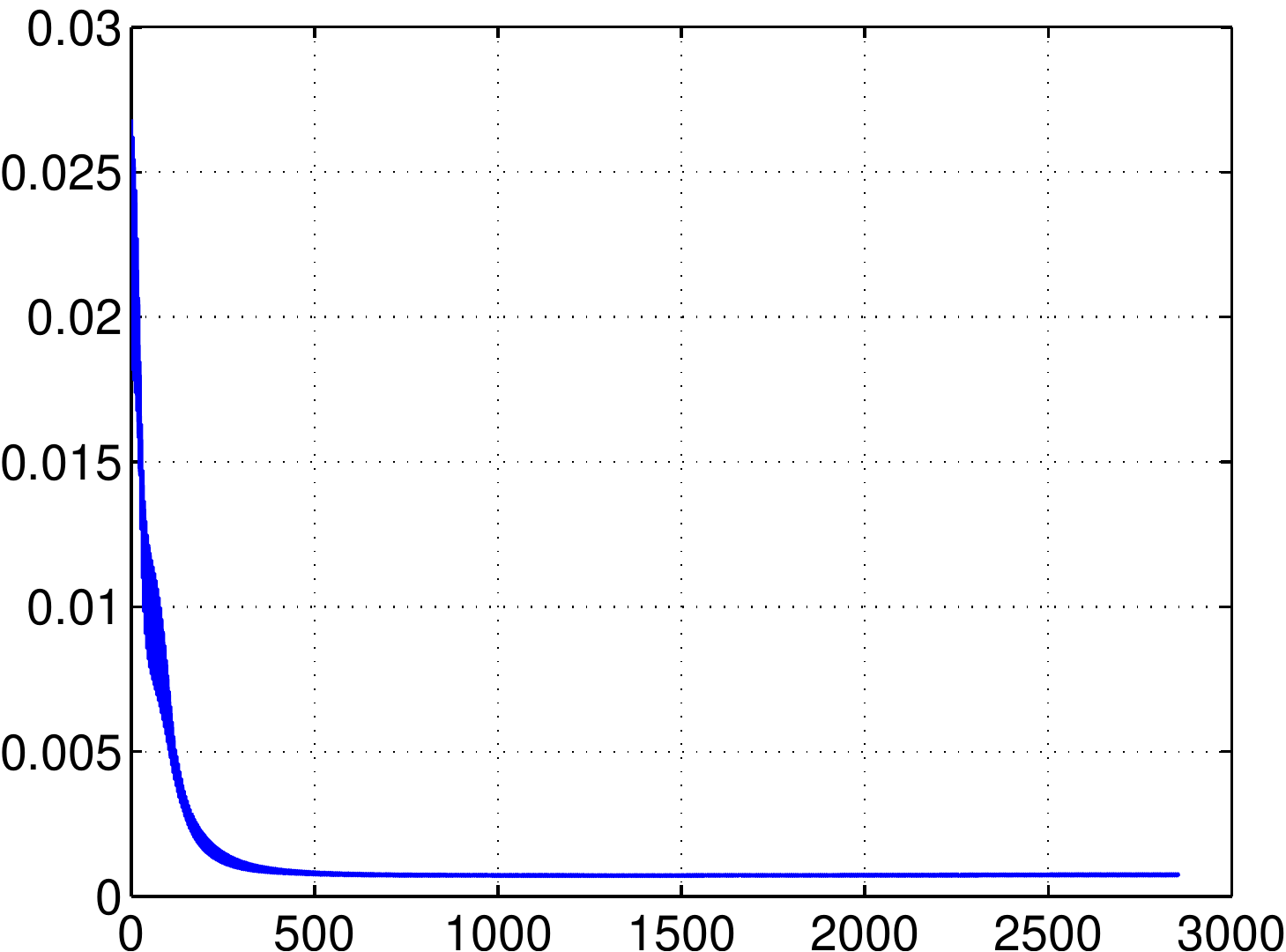}}
\subfigure[PID]{\label{fig.iris}\includegraphics[width=1.3in]{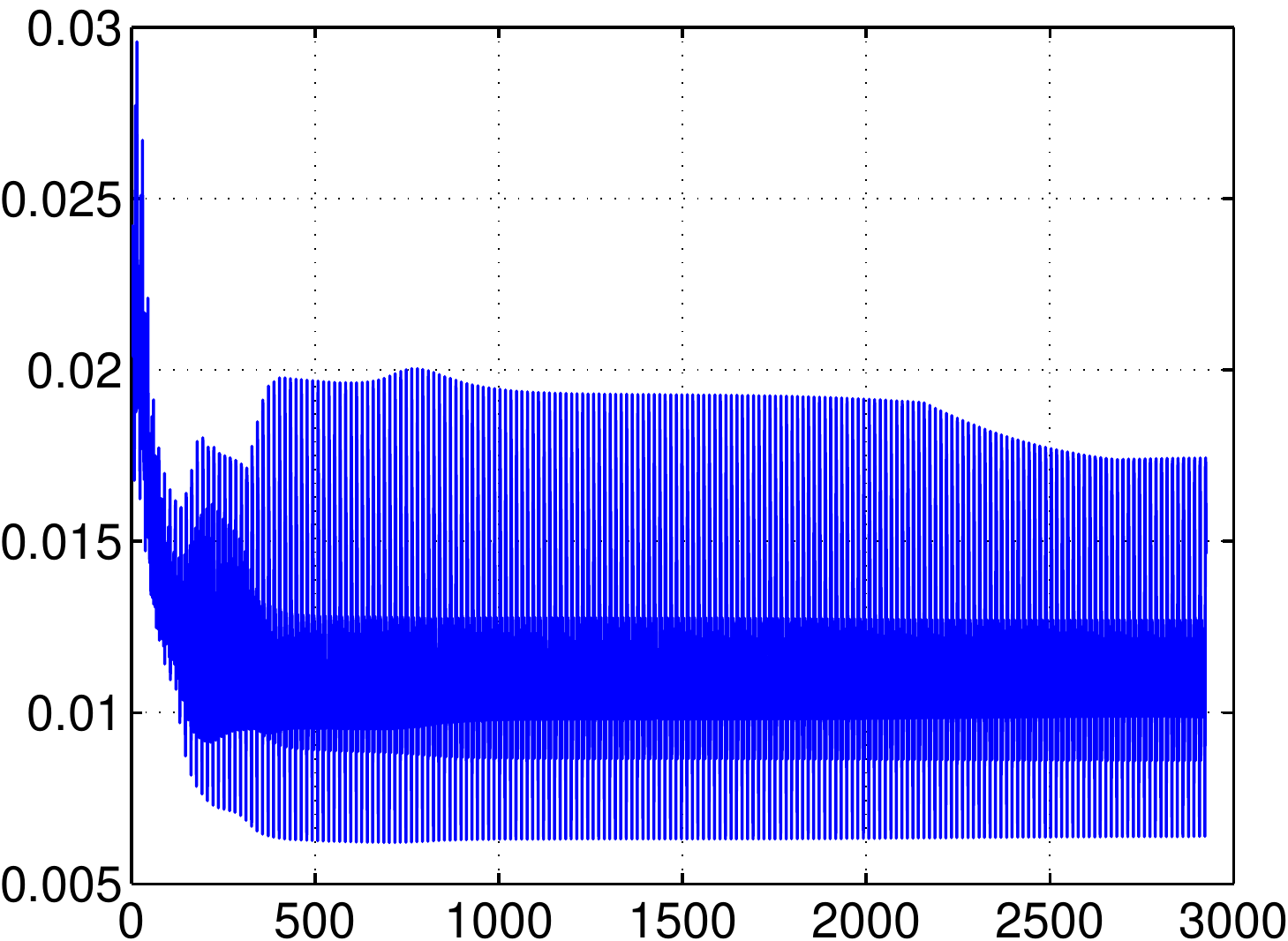}}
%\subfigure[Yeast]{\label{fig.tae}\includegraphics[width=1.3in]{yeast_8_4}}
\subfigure[BC]{\label{fig.iris}\includegraphics[width=1.3in]{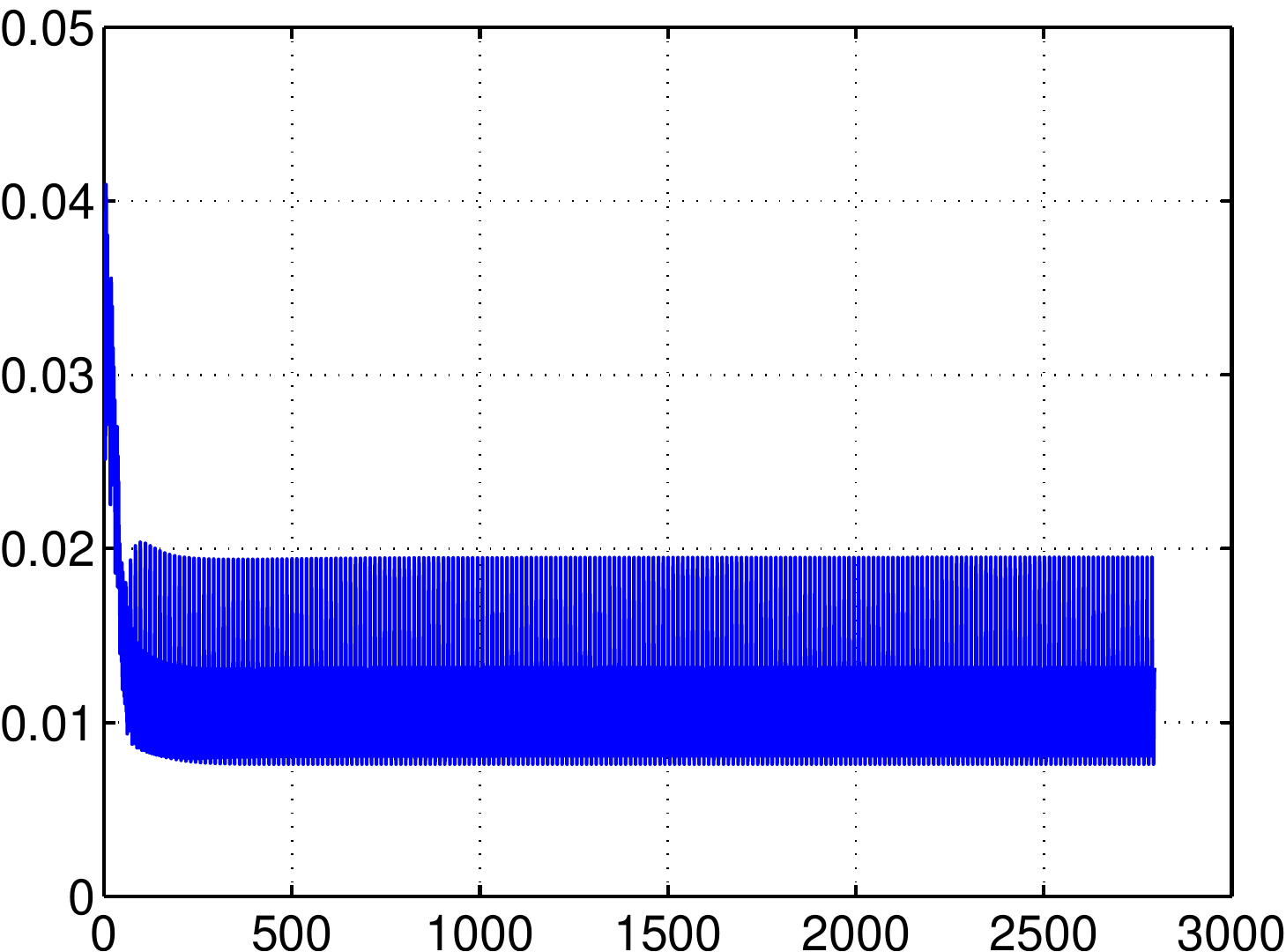}}
\subfigure[Glass]{\label{fig.tae}\includegraphics[width=1.3in]{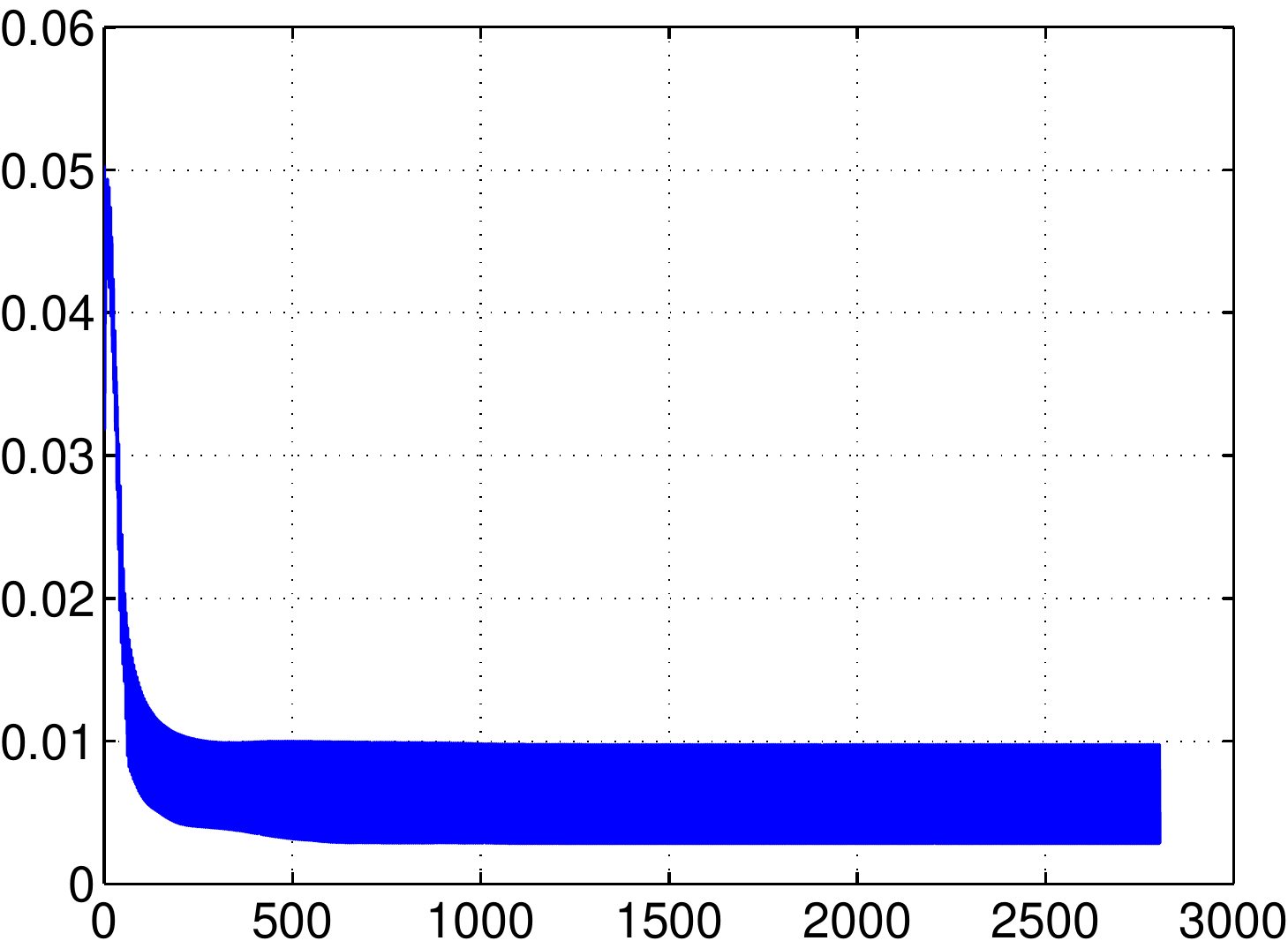}}
\subfigure[Wine]{\label{fig.iris}\includegraphics[width=1.3in]{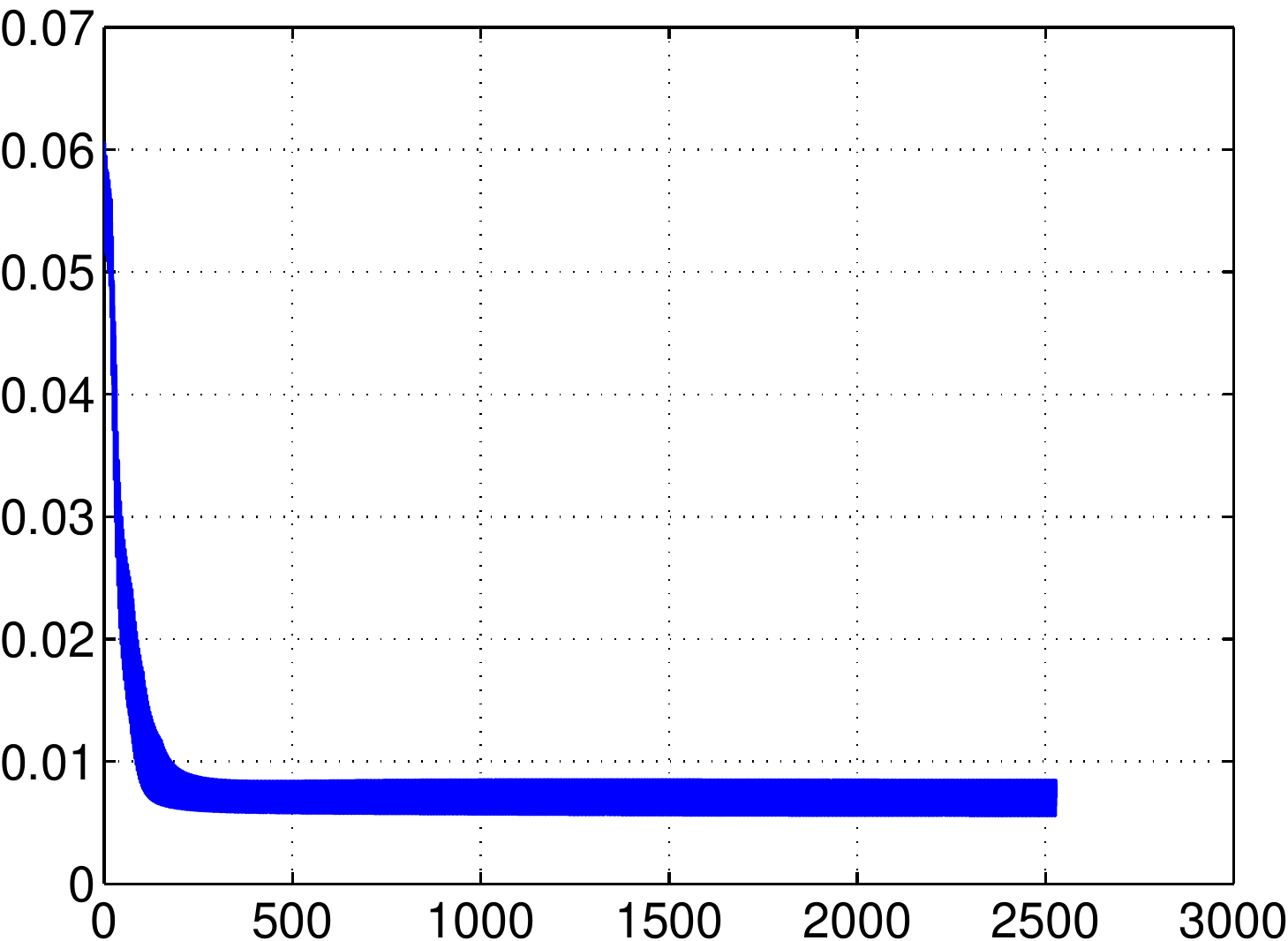}}
\subfigure[Zoo]{\label{fig.tae}\includegraphics[width=1.3in]{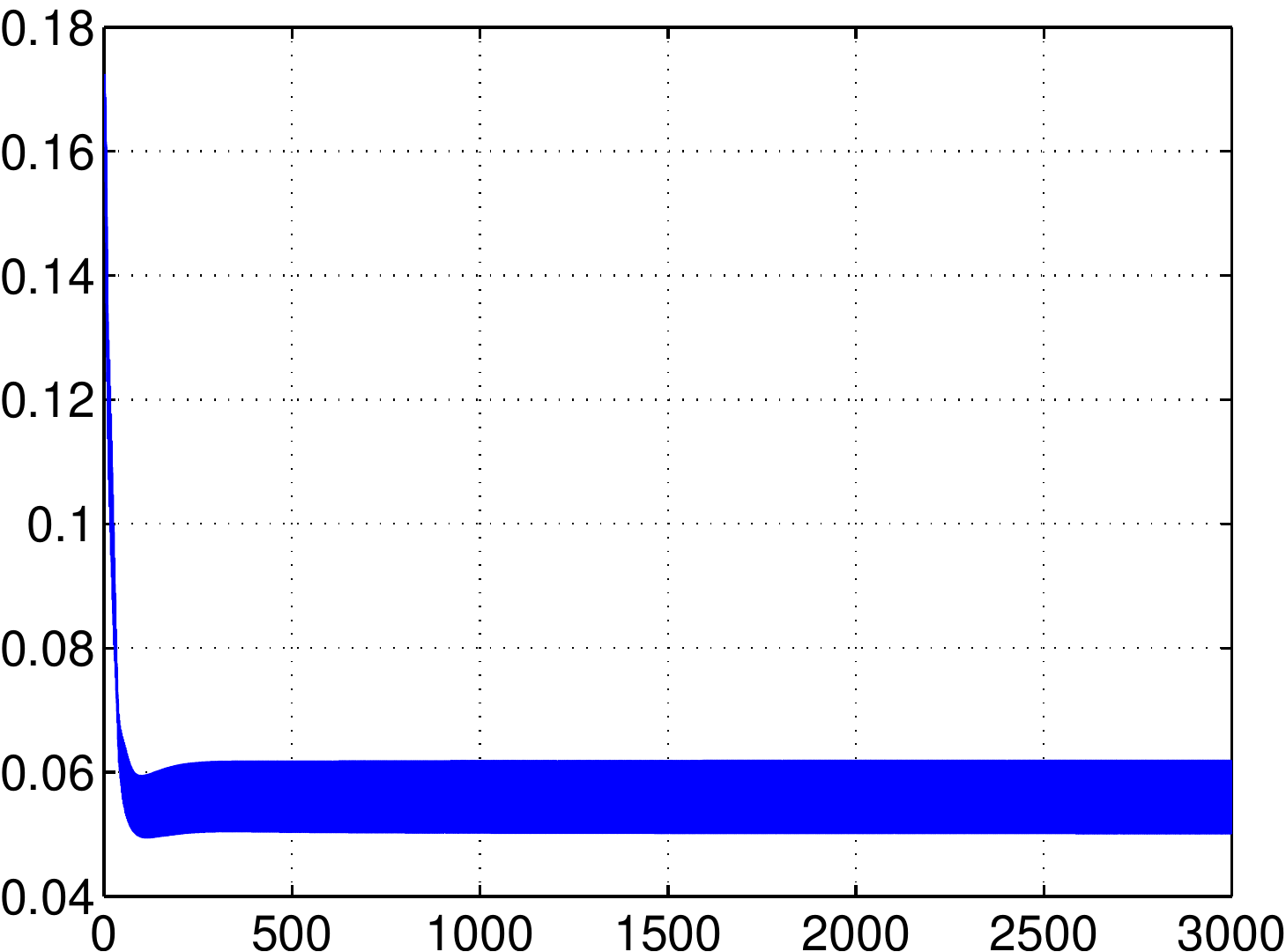}}
\subfigure[Parkinsons]{\label{fig.iris}\includegraphics[width=1.3in]{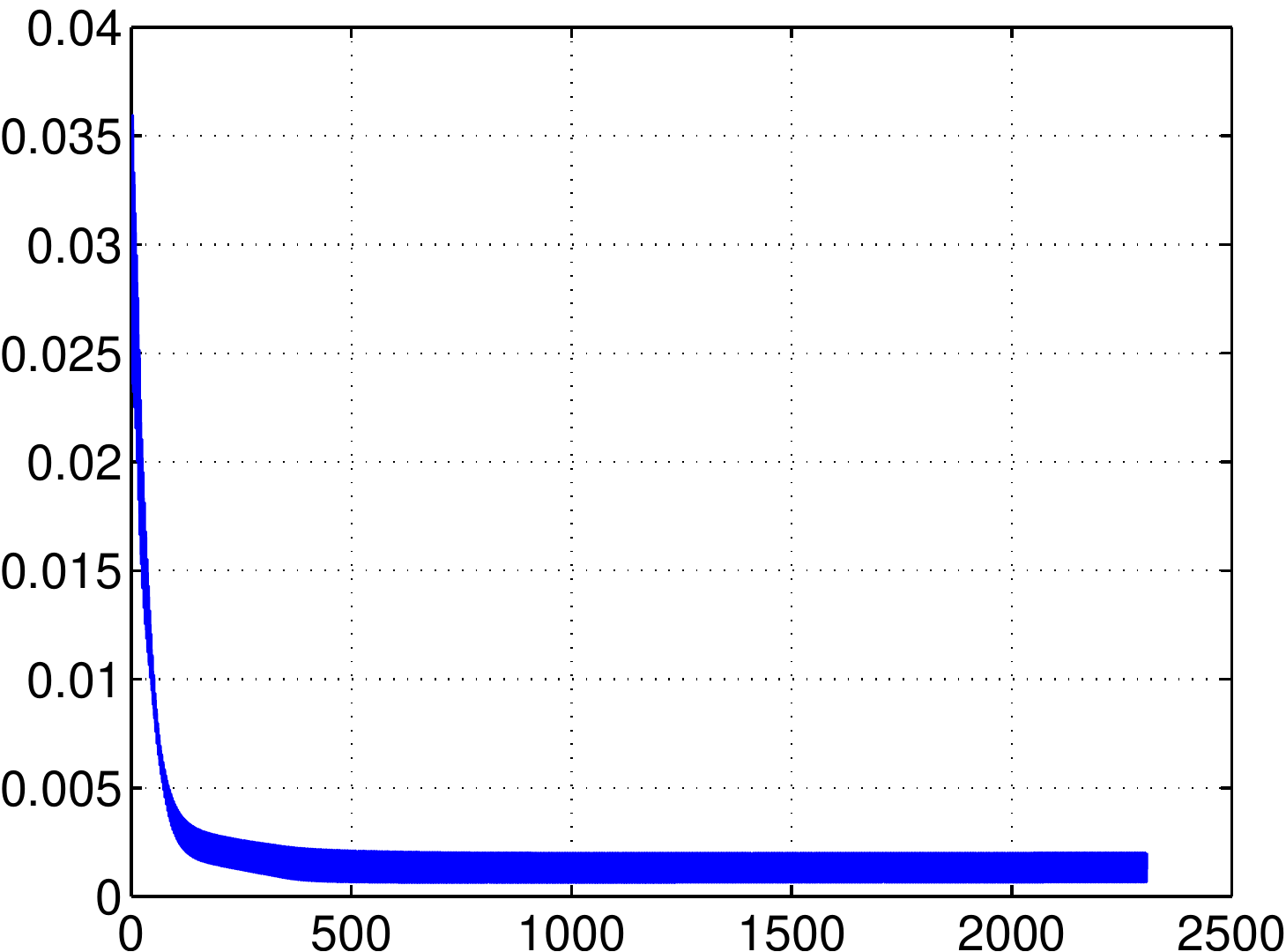}}
\subfigure[WFRN]{\label{fig.tae}\includegraphics[width=1.3in]{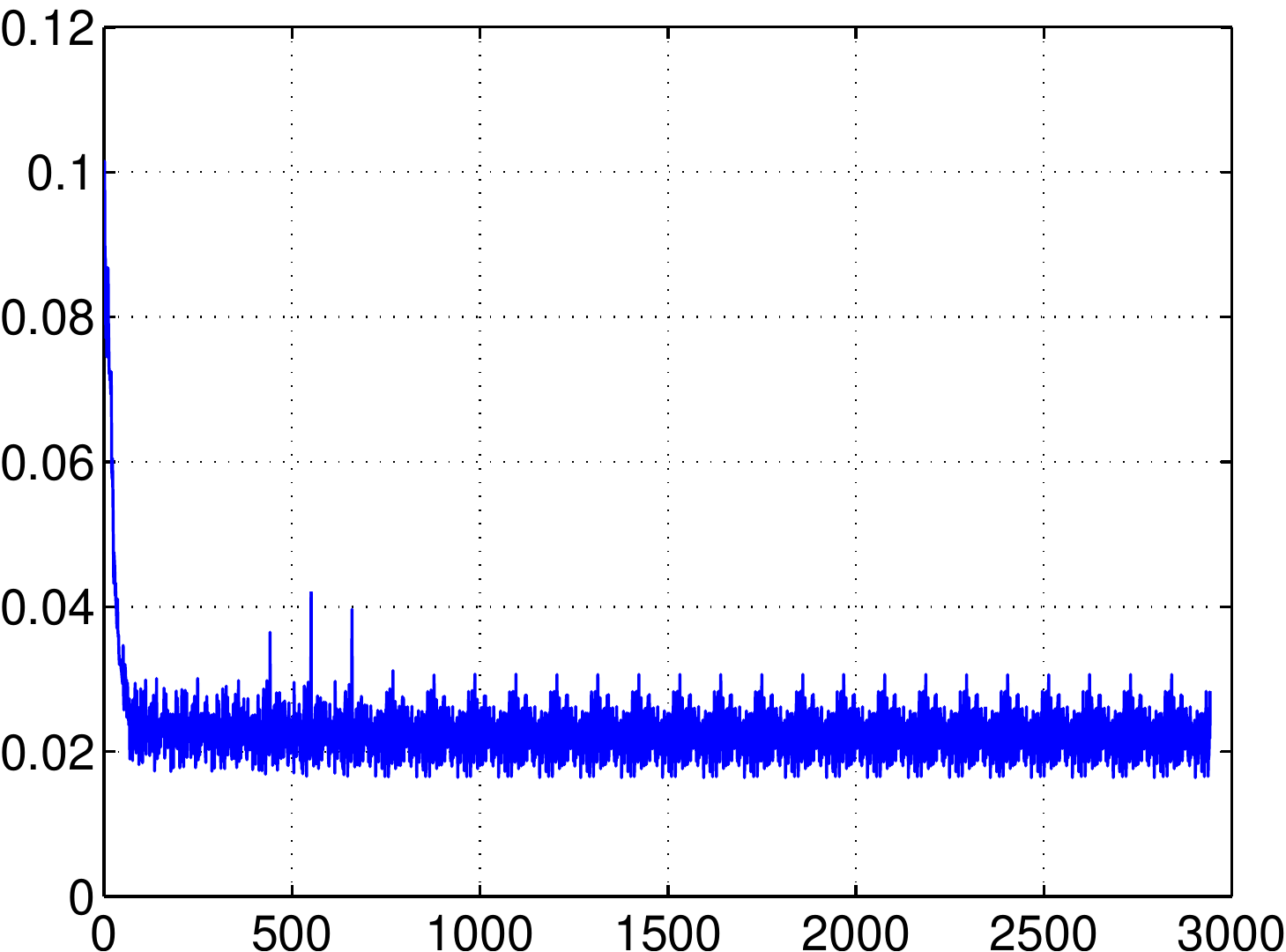}}
\subfigure[Ionosphere]{\label{fig.iris}\includegraphics[width=1.3in]{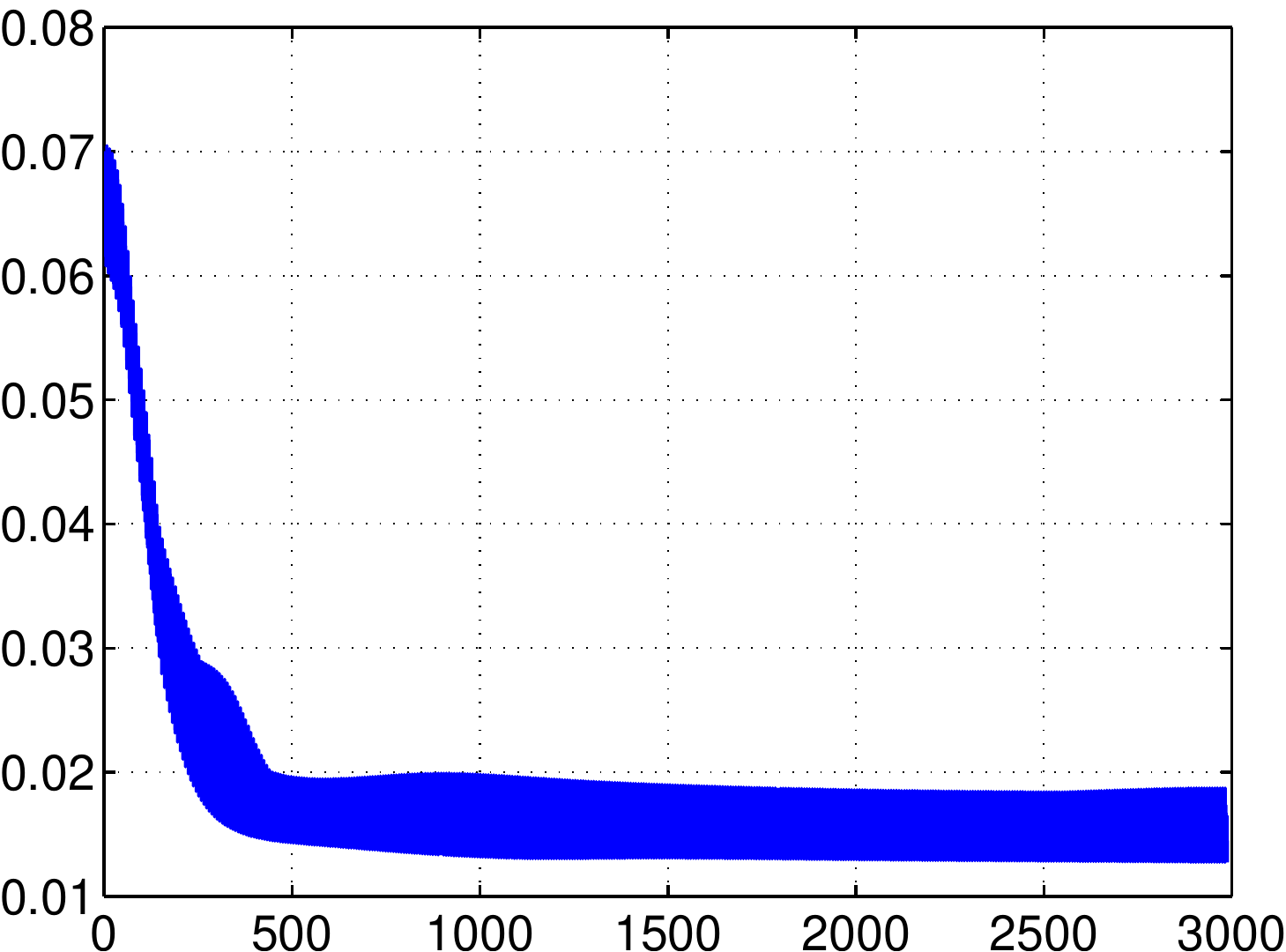}}
\subfigure[SL]{\label{fig.tae}\includegraphics[width=1.3in]{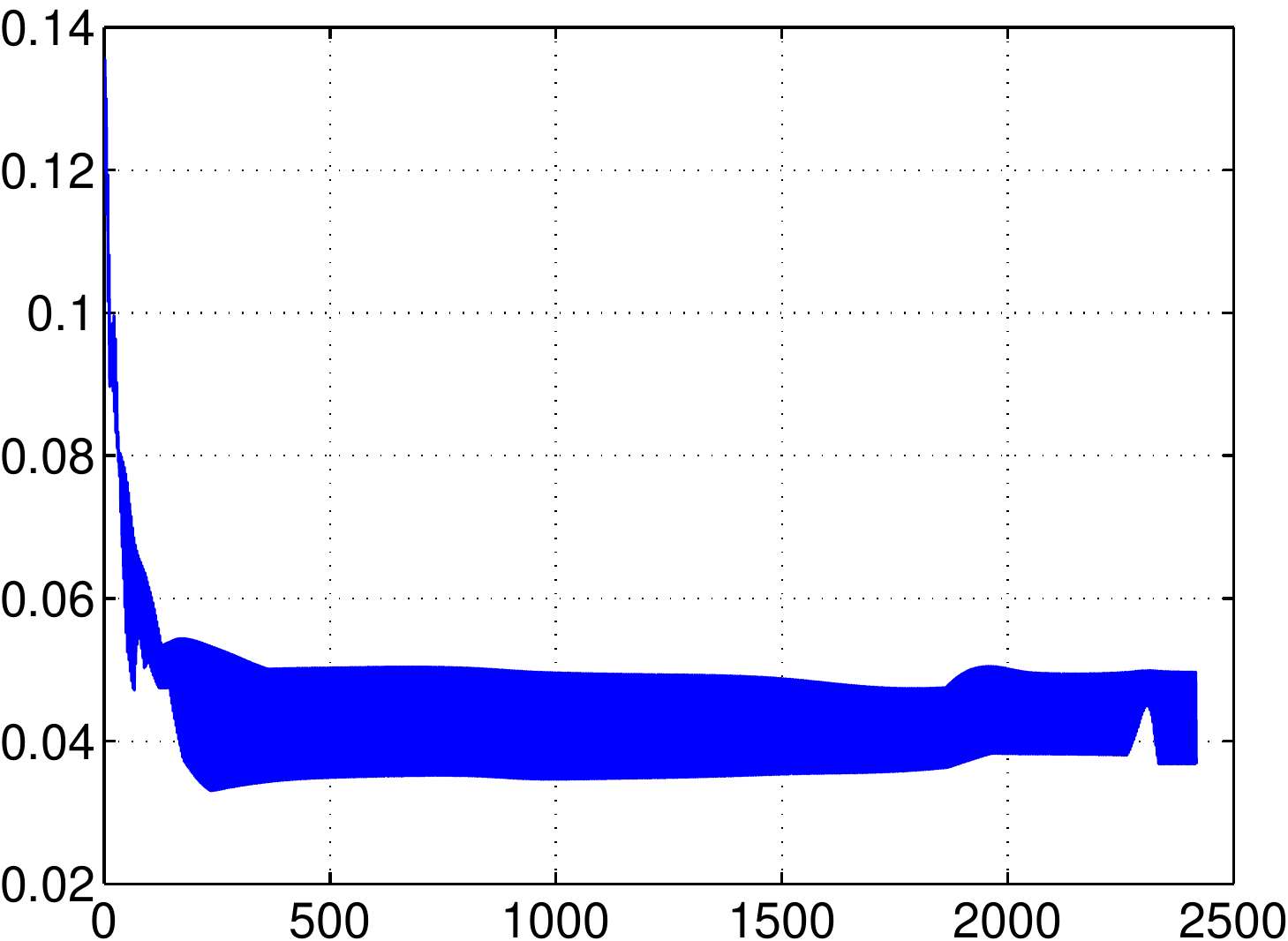}}
\subfigure[FOTP]{\label{fig.iris}\includegraphics[width=1.3in]{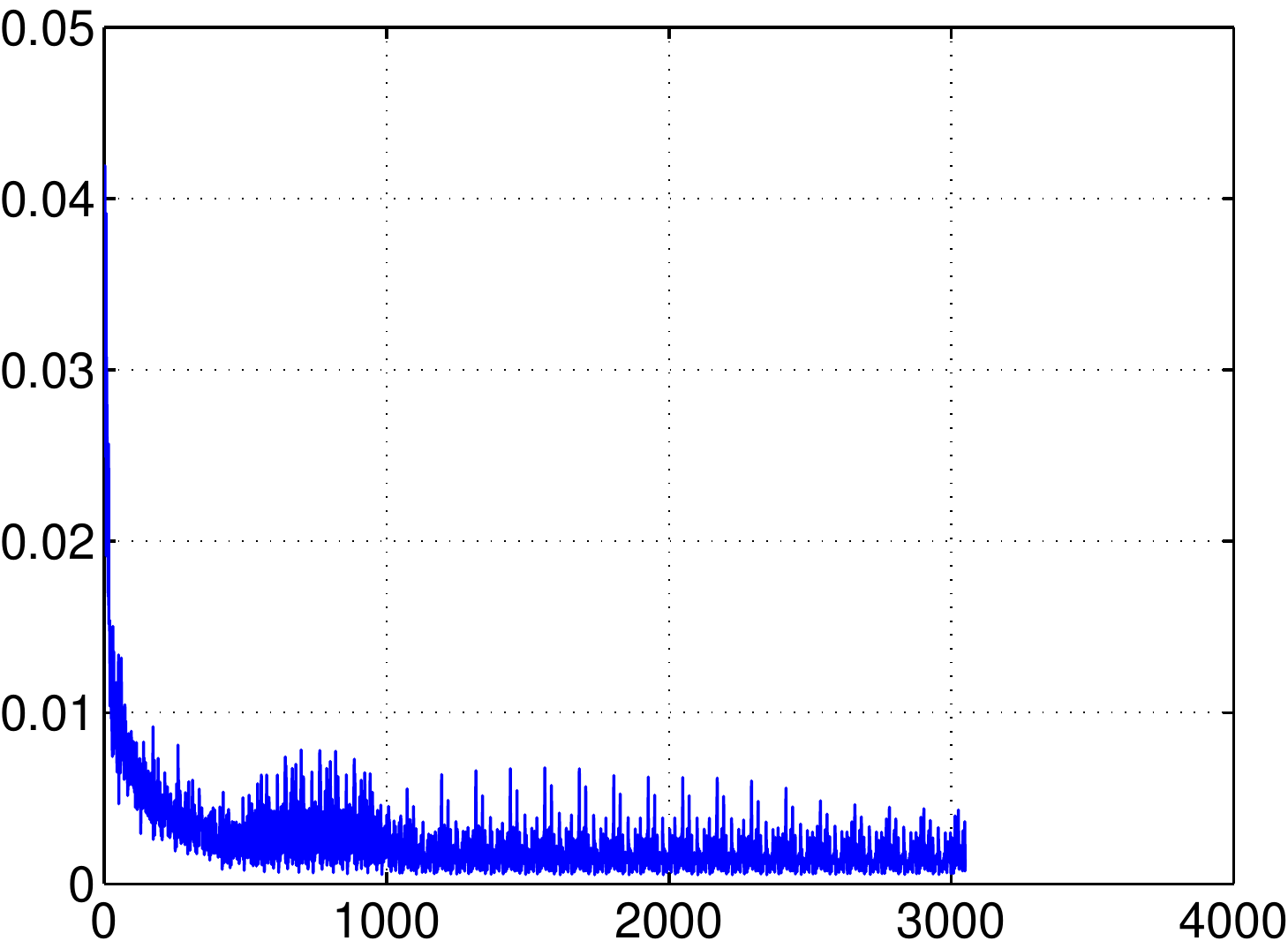}}
\subfigure[Sonar]{\label{fig.tae}\includegraphics[width=1.3in]{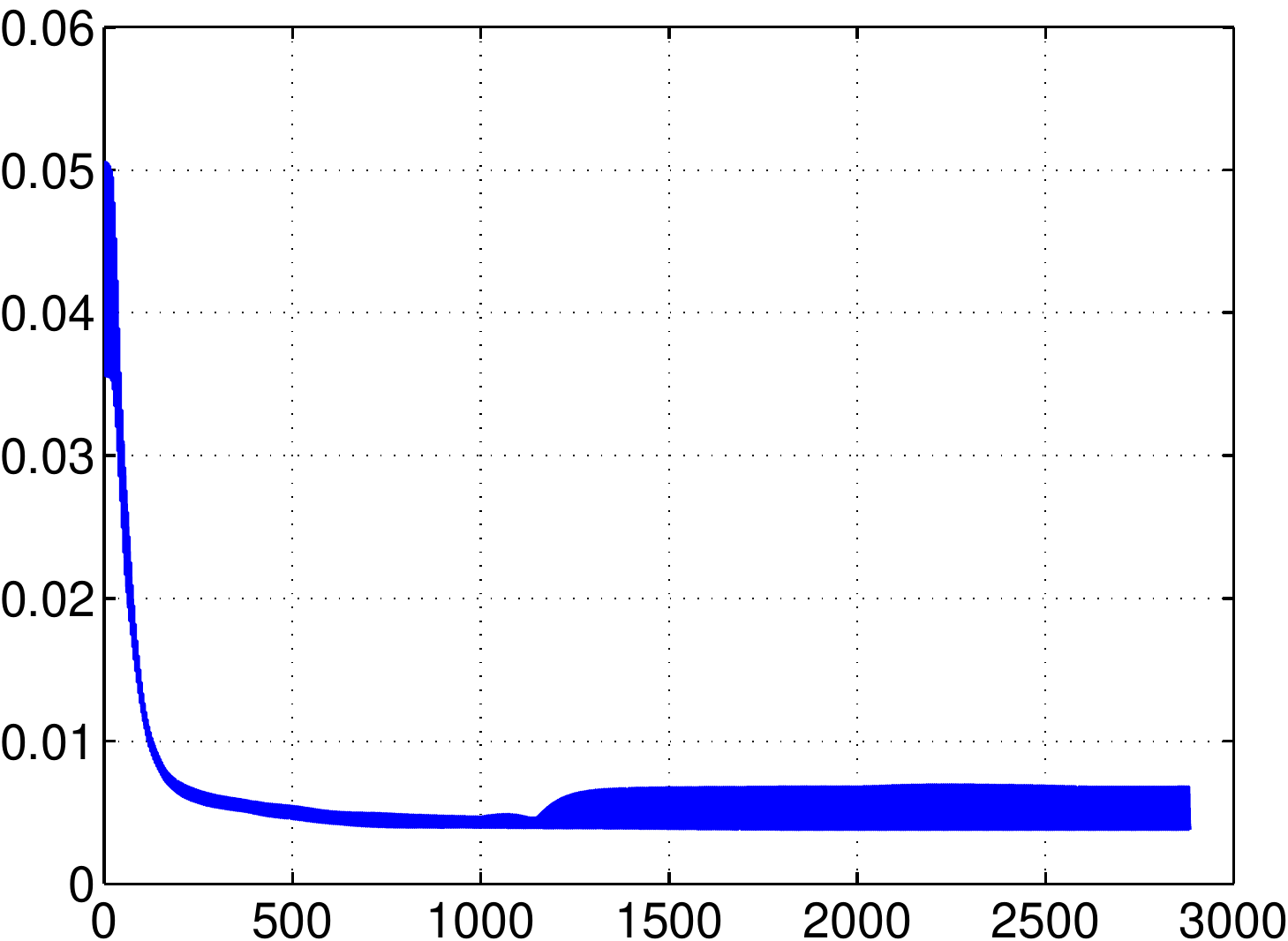}}
\subfigure[CA]{\label{fig.tae}\includegraphics[width=1.3in]{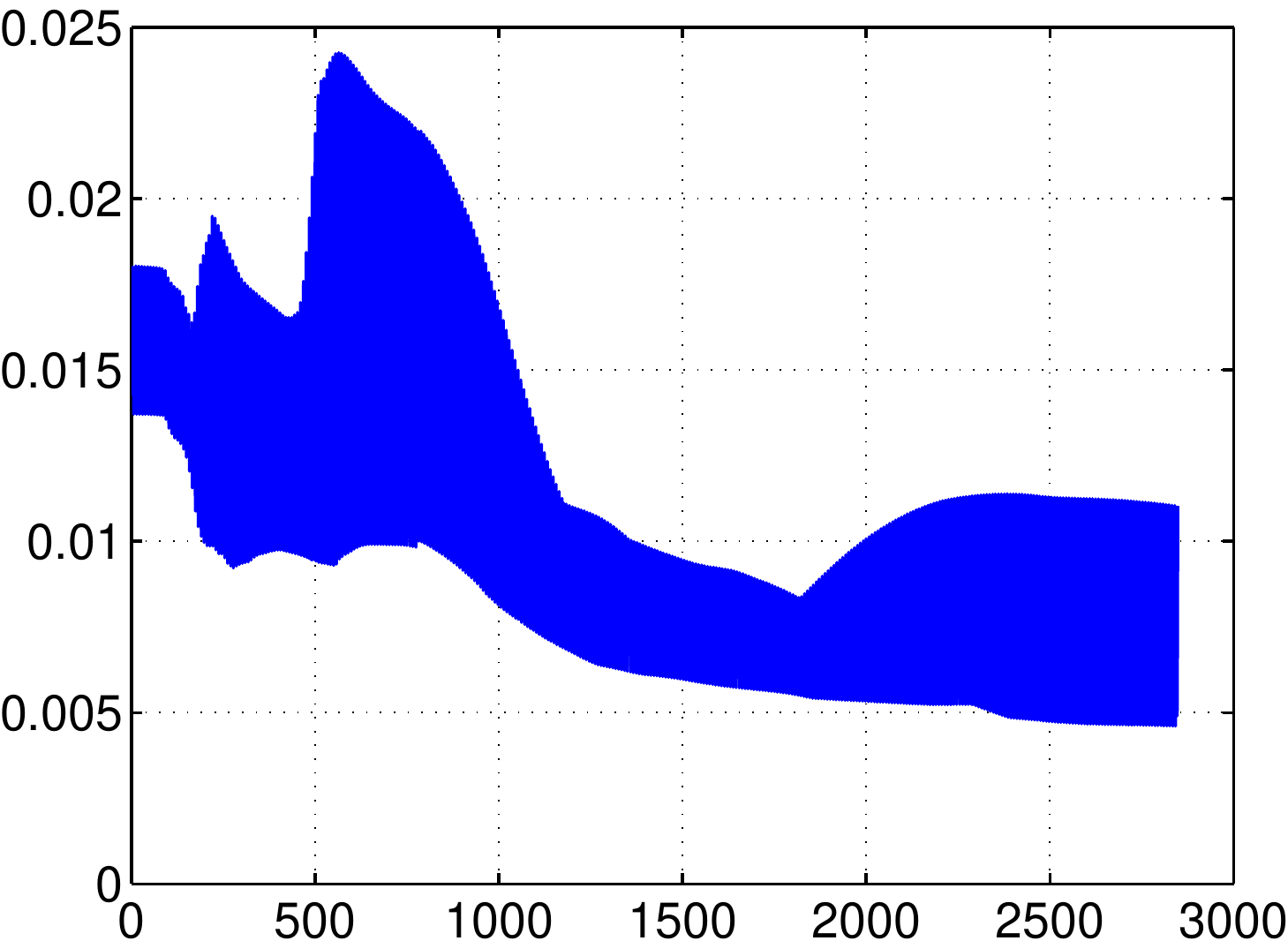}}
\caption{Reconstruction error (Y-axis) versus iteration number (X-axis) of nonnegative LRNN autoencoders for UCI real-world datasets.} \label{fig.uci}
\end{figure*}

{\bf Results of UCI real-world datasets:} Let $N$ denote the attribute number in a dataset. The structures of the LRNN autoencoders used are $N \rightarrow \text{round}(N/2)$,
%$N \rightarrow \text{round}(N/2) \rightarrow \text{round}(N/4)$,
where the operation $\text{round}(\cdot)$ produces the nearest integer number of the element.
The attribute values are linear normalized in range $[0,1]$. The size of mini-batches is set as 50 for all datasets.
Curves of reconstruction errors versus iterations are given in Figure \ref{fig.uci}.
We see that the reconstruction errors generally decrease as the number of iterations increases.
These results also demonstrate the efficacy of the nonnegative LRNN autoencoders equipped with the training algorithms.

\section{Simulating the spiking random neural network}
The advantage of a spiking model, such as the LRNN autoencoder, lays on its highly-distributed nature.
In this section, rather than numerical calculation, we simulate the stochastic spiking behaviors of the LRNN autoencoder. The simulation in this section is based on the numerical experiment of Subsection \ref{sec.mnist_results}.
Specifically, in Subsection \ref{sec.mnist_results}, we construct a LRNN autoencoder of structure $784 \rightarrow 100$ (with appropriate weights found), which has three layers: the visual layer (784 neurons), hidden layer (100 neurons) and output layer (784 neurons). First, an image with $28 \times 28 = 784$ attributes is taken from the MNIST dataset. Each visual neuron receives excitatory spikes from outside the network in a Poisson stream with the rate being the corresponding attribute value in the image.
When activated, the visual neurons fire excitatory spikes to the hidden neurons according to the Poisson process with rate 1 (meaning $w_{v,h}=p^+_{v,h}$). When the $v$th visual neuron fires to the hidden layer, the spike goes to the $h$th hidden neuron with probability $p^+_{v,h}$ or it goes outside the network with probability $1-\sum_{h=1}^{H}p^+_{v,h}$. The hidden neurons fire excitatory spikes to the output layer in a similar manner subjecting to $\overline{w}_{h,o}$. The firing rate of output neurons is $1$ and the spikes go outside the network with probability 1.

In the simulation, we call it an event whenever a spike gets in from outside the network or a neuron fires. During the simulation, we observe the potential (the level of activation) of each neuron once every 1,000 events. Let $k_{i,b}$ represent the $b$th observation of the $i$th neuron. We estimate the average potential of the $i$th neuron, denoted by $\bar{k}_i$, simply by averaging observations, i.e., $\bar{k}_i \approx (\sum_{b=1}^{B}k_{i,b})/B$. Let $q_i$ denote the probability that the $i$th neuron is activated. The relation between $q_i$ and $\bar{k}_i$ is known as $\bar{k}_i=q_i/(1-q_i)$. Then, the value of $q_i$ can be estimated during the simulation as:
\begin{equation}
q_i =  \frac{\bar{k}_i}{1+\bar{k}_i} \approx \frac{(\sum_{b=1}^{B}k_{i,b})/B}{1+(\sum_{b=1}^{B}k_{i,b})/B}.
\end{equation}

\begin{figure}\centering
\psfrag{ori}[c][c][0.7]{Numerical}%
\psfrag{10000}[c][c][0.7]{After 10000 events}%
\psfrag{100000}[c][c][0.7]{After 100000 events}%
\psfrag{1000000}[c][c][0.7]{After 1000000 events}%
\includegraphics[width=4.8in]{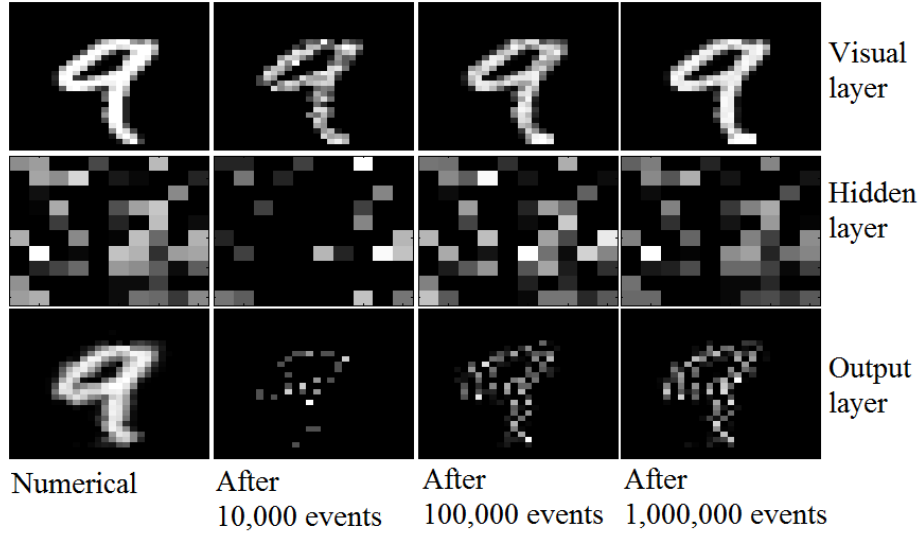}
\caption{Comparisons between numerical and spiking–behavior simulation results of difference layers in a LRNN autoencoder.} \label{fig.lrnn_simu}
\end{figure}

In Figure \ref{fig.lrnn_simu}, we visualize the estimated values of $q_i$ for all neurons in different layers after 10,000, 100,000 and 1,000,000 events during the simulation. For comparison, numerical results from Subsection \ref{sec.mnist_results} are also given in Figure \ref{fig.lrnn_simu}.
At the beginning, simulation results of only the visual layer are close to its numerical results.
As time evolves, the simulation results of the hidden and output layers and their corresponding numerical results become more and more similar.
These results demonstrate that the LRNN autoencoders have the potential to be implemented in a highly distributed and parallel manner.

%\begin{figure}\centering
%\psfrag{ori}[c][c][0.7]{Numerical}%
%\psfrag{10000}[c][c][0.7]{After 10000 events}%
%\psfrag{100000}[c][c][0.7]{After 100000 events}%
%\psfrag{1000000}[c][c][0.7]{After 1000000 events}%
%\subfigure[Visual layer]{\label{fig.visual}\includegraphics[width=3in]{5MNIST_784_100_in_all.png}}
%\subfigure[Hidden layer]{\label{fig.hidden}\includegraphics[width=3in]{5MNIST_784_100_hidd_all}}
%\subfigure[Output layer]{\label{fig.hidden}\includegraphics[width=3in]{5MNIST_784_100_out_all}}
%\caption{Comparisons between numerical and simulation results of difference layers in a shallow LRNN autoencoder.} \label{fig.lrnn_simu}
%\end{figure}

\section{Conclusions}
New nonnegative autoencoders (the shallow and multi-layer LRNN autoencoders) have been proposed based on the spiking RNN model, which adopt the feed-forword multi-layer network architecture in the deep-learning area. To comply the RNN constraints of nonnegativity and that the sum of probabilities is no larger than 1, learning algorithms have been developed by adapting weight update rules from the NMF area.
Numerical results based on typical image datasets including the MNIST, Yale face and CIFAR-10 datesets and 16 real-world datasets from different areas have well verified the robust convergence and reconstruction performance of the LRNN autoencoder. In addition to numerical experiments, we have conducted simulations of the autoencoder where the stochastic spiking behaviors are simulated. Simulation results conform well with the corresponding numerical results. This demonstrates that the LRNN autoencoder can be implemented in a highly distributed and parallel manner.

\bibliographystyle{IEEEtran}
\bibliography{RNN}

\end{document}